\documentclass[journal]{IEEEtran}

\ifCLASSINFOpdf
\else
\fi

\usepackage{subfig}
\usepackage{graphicx} 
\usepackage[noend]{algorithm2e}
\usepackage{algorithmic}
\usepackage{setspace}
\usepackage{url}
\usepackage{amsmath,amsthm}
\usepackage{amssymb,wasysym}
\usepackage{mathrsfs}
\usepackage{cite}
\usepackage{xcolor}
\usepackage{wrapfig}
\usepackage{verbatim}
\usepackage{dsfont}
\usepackage{multirow}
\usepackage{breqn,xspace}
\usepackage{bbm}

\ifodd 0
\newcommand{\rev}[1]{{\color{blue}#1}} 
\newcommand{\newrev}[1]{{\color{red}#1}} 

\else
\newcommand{\rev}[1]{#1}
\newcommand{\newrev}[1]{#1} 
\fi

\newcommand{\name}{LEO-Split\xspace}

\begin{document}

\title{\name: A Semi-Supervised Split Learning Framework over LEO Satellite Networks}

\author{Zheng Lin, Yuxin Zhang, Zhe Chen,~\IEEEmembership{Member,~IEEE}, Zihan Fang, Cong Wu,~\IEEEmembership{Member,~IEEE},  Xianhao Chen,~\IEEEmembership{Member,~IEEE}, Yue Gao,~\IEEEmembership{Fellow,~IEEE}, and Jun Luo,~\IEEEmembership{Fellow,~IEEE}

\thanks{Z. Lin, Y. Zhang, Z. Chen and Y. Gao are with the Institute of Space Internet, Fudan University, Shanghai 200438, China, and the School of Computer Science, Fudan University, Shanghai 200438, China (e-mail: zlin20@fudan.edu.cn; yxzhang24@m.fudan.edu.cn; zhechen@fudan.edu.cn; gao.yue@fudan.edu.cn). Z. Lin is also with the Department of Electrical and Electronic Engineering, University of Hong Kong, Pok Fu Lam, Hong Kong, China.}
\thanks{Z. Fang is with the Department of Computer Science, City University of Hong Kong, Kowloon, Hong Kong SAR, China (e-mail: zihanfang3-c@my.cityu.edu.hk).}
\thanks{C. Wu and X. Chen are with the Department of Electrical and Electronic Engineering, University of Hong Kong, Pok Fu Lam, Hong Kong, China (e-mail: congwu@hku.hk; xchen@eee.hku.hk).}
\thanks{Jun Luo is with the School of Computer Science and Engineering, Nanyang Technological University, Singapore (e-mail: junluo@ntu.edu.sg).}
\thanks{\textit{(Corresponding author: Zhe Chen; Yue Gao)}}
}


\maketitle

\begin{abstract}
%
%
Recently, the increasing deployment of LEO satellite systems has enabled various space analytics (e.g., crop and climate monitoring), \newrev{which heavily relies on
the advancements} in deep learning (DL). However, 
the intermittent connectivity between LEO satellites and ground station (GS) significantly hinders the timely transmission of raw data to GS for centralized learning,
while the scaled-up DL models hamper distributed learning
on resource-constrained LEO satellites.
Though \textit{split learning} (SL) can be \newrev{a potential solution} 
to these problems by partitioning a model and offloading primary training workload to GS,
the labor-intensive 
labeling process remains \newrev{an obstacle,
%
%
with intermittent connectivity and data heterogeneity 
being other challenges.}
In this paper, we propose \name, a \textit{semi-supervised} (SS) SL \newrev{design} tailored for satellite networks to {combat} these challenges. \newrev{Leveraging SS learning to handle (labeled) data scarcity, we} construct an auxiliary model to tackle the training failure of the satellite-GS non-contact time.
Moreover, we propose a pseudo-labeling algorithm to rectify data imbalances across satellites.
Lastly, an adaptive activation interpolation scheme is devised to prevent the overfitting of {server-side} sub-model training at GS. 
Extensive experiments with real-world LEO satellite {traces}~(e.g., Starlink) demonstrate that our \name framework achieves superior performance compared to state-of-the-art benchmarks.
\end{abstract}


\IEEEpeerreviewmaketitle

\section{Introduction}\label{sec:introduction} 

\IEEEPARstart{R}{ecently}, more and more low earth orbit~(LEO) satellites have been sent into space to build satellite-enabled broadband Internet and sensing networks{~\cite{liu2024democratizing,yuan2024satsense,zhao2024leo,planet_lab}}. Specifically, to build the LEO satellite meta-constellation, commonly known as Starlink, SpaceX plans to launch approximately 42,000 LEO satellites~\cite{ahmmed2022digital,lin2024fedsn,zhang2024satfed}. 
Beyond providing enhanced broadband internet connectivity~\cite{ZW_VTC_2021},
these LEO satellites are equipped with multimodal sensors capable of collecting a wide spectrum of informative sensor data, typically including spectral data~{\cite{singh2024spectrumize,peng2024sums}} and terrestrial images{~\cite{singh2022selfiestick,huang2024d}.} \newrev{More importantly, LEO satellites suggest a new type of infrastructure for networking and sensing in the space, and multiple tenants~(e.g., mobile operators and remote sensing companies) may build their services based on this infrastructure~{\cite{liu2024democratizing,yuan2023graph}}. To fully take advantage of the rich information provided by various sensors of LEO satellites, these} tenants can leverage deep learning~(DL) to resolve many global challenges such as food safety~\cite{aragon2018cubesats}, disease spread~\cite{franch2020spatial}, and climate change~\cite{shukla2021enhancing}. 


However, deploying DL frameworks over satellite networks still faces significant obstacles. Firstly, the excessive costs in both manpower (experts with interdisciplinary expertise) and time
render the manual annotation of vast satellite datasets impractical. Consequently, sensing data from LEO satellites is commonly unlabeled, substantially deteriorating data utilization and analysis efficacy. Secondly, the conventional centralized DL paradigm entails raw data gathering and processing at a central server~\cite{lin2023efficient,10175391}. In satellite networks, due to constrained power, size, and mass of a satellite, the computing power is often less than a central server~(e.g., LEO satellite of Planet Labs equipped with common NVIDIA Jetson~\cite{jetson} has only 1/10
of RTX 3090's GPU cores), \newrev{and it is shared among multiple tenants. Meanwhile, raw data sharing among different tenants is hindered by regulation restrictions, commercial interests, and data ownership concerns~\cite{coffer2020balancing,Monmonier_2004,zhang2024fedac,hu2024accelerating}}.
Last but not least, though federated learning (FL) \cite{mcmahan2017communication,zw2024specbreathing} offers a potential solution to these challenges, training large neural network models on resource-constrained LEO satellites remains a prohibitive task as DL models scale up{~\cite{han2023splitgp,li2021hermes,zhang2024lolafl,lin2024adaptsfl}}.

To tackle these challenges, a straightforward thought could be to combine  \textit{split learning}~(SL)~\cite{vepakomma2018split,lin2024split} with semi-supervised learning~(SSL) for LEO satellite networks. On the one hand, the SL framework partitions a neural network model into client-side and server-side sub-models, placing them in satellites and \newrev{ground station} (GS) to train together. \newrev{Since the size of a client-side sub-model is much smaller than that of a server-side one, SL can substantially reduce the training load on satellites~\cite{lin2024split,lin2024hierarchical}.}
After training, the GS aggregates multiple versions of the server-side and client-side sub-models for the next training round. On the other hand, SSL can be built on top of SL to reduce extensive labeling labor. SSL leverages a small number of labeled data, and large amounts of unlabeled data to train a neural network model. State-of-the-art SSL frameworks predominantly rely on pre-defined fixed pseudo-label thresholds for unlabeled data to guide model training~\cite{sohn2020fixmatch,berthelot2019mixmatch,berthelot2020remixmatch,xie2020unsupervised}. 

\begin{figure}[t]
\centering
\includegraphics[width=.86\columnwidth]{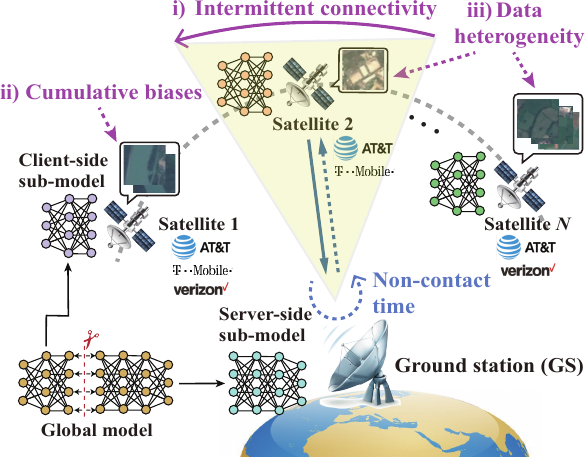}
\caption{
\newrev{Though SL may facilitate efficient satellite-GS collaborative model training, i) satellite-GS intermittent connectivity, ii) cumulative biases, and iii) data heterogeneity still pose significant challenges to real-world implementations.}}
\label{fig:sl_over_satellite_network}
\end{figure}

However, 
directly combining SL with SSL faces three major practical problems over LEO satellite networks, as shown in Fig.~\ref{fig:sl_over_satellite_network}. Firstly,
conventional SL framework relies heavily on continuous connectivity between satellites and GS to exchange data
for model training~\cite{vepakomma2018split}, but contact time between satellites and GS typically occupies only a small fraction of its orbital period {(e.g., 4-minute contact time in 95-minute orbital period~\cite{starlink_gp})}, resulting in training failure for SL.
Secondly, \newrev{even if vanilla SSL may have reduced demand on the labeled data, it inevitably brings model biases caused by the lack of full-fledged training data. Unfortunately, the tolerance to such biases can be significantly lowered due to SL's model aggregation process that accumulates such biases. Consequently, naively combining SL with SSL could lead to
poor model generalization and increased risk of overfitting.}
Lastly, discrepancies in orbital circumstances, storage, and sensing capabilities~(e.g., image resolution) lead to data heterogeneity~(i.e., data classes and quantity imbalances across local datasets). This inevitably causes catastrophic forgetting~\cite{huang2022learn,qu2022rethinking} and also renders model bias severer~\cite{li2023revisiting,abay2020mitigating}, thus further deteriorating model generalization.
{To tackle the above challenges, we propose \newrev{\name as a novel} {s}emi-{s}upervised \underline{split} {l}earning framework specifically tailored for \underline{LEO} satellite networks. First, in order to resolve the training failure \newrev{during the satellite-GS non-contact time,} we construct an auxiliary model for each satellite; \newrev{it acts as a light surrogate to}
the whole model to enable independent training of the \newrev{satellite's}
sub-model. Second, \newrev{to combat data heterogeneity and poor generalization caused by cumulative biases,} we adaptively adjust the pseudo-labeling threshold to \newrev{balance} the data distribution among all satellites during model training. Finally, to resolve the conflict between 
overfitting caused by limited activations 
during short contact time and continuous data updates on satellites, we adaptively interpolate activations of labeled and high-confidence pseudo-labeled data to mitigate performance degradation brought by overfitting.
%
%
The key contributions of this paper are summarized as follows:


\begin{itemize}
  \item To the best of our knowledge, \name is the first semi-supervised SL framework specifically designed for LEO satellite networks.  
  %
  %
  \item We construct an auxiliary model for enabling independent updates of the client-side sub-model, aiming to resolve the training failure during the satellite-GS non-contact time.
  \item We design a pseudo-labeling algorithm at satellite to enable
  data-balanced model training, largely alleviating the data heterogeneity across satellites.
  \item We devise an adaptive activation interpolation strategy to compensate the data richness gap between satellites and GS, thus eliminating the harmful effects induced by server-side overfitting.
  %
  \item We empirically evaluate \name with extensive experiments. The results demonstrate that \name outperforms state-of-the-art frameworks.
\end{itemize}

The rest of the paper is organized as follows.
Section~\ref{sec:background_motivation} motivates the design of \name by revealing the challenges in current LEO satellite networks.
Sec.~\ref{sec:s3l_frame} presents the system design of \name.
Sec.~\ref{sec:implementation} introduces the system implementation, followed by performance evaluation in Sec.~\ref{sec:simulation results}.
Related works and technical limitations are discussed in Sec.~\ref{sec:related_work}.
Finally, conclusions are presented in Sec.~\ref{sec:conclusion}.

\section{Challenges and Motivation}  \label{sec:background_motivation}

In this section, we investigate the impacts of i) intermittent connectivity, ii) SSL cumulative biases caused by SL model aggregation, and iii) data heterogeneity among different satellites on SL performance. 


\subsection{Intermittent and Limited Connectivity }\label{subsec:limit_contact}
Conventional SL~\cite{vepakomma2018split,lin2023efficient} trains client-side and server-side sub-models separately, thus demanding continuous \textit{smashed data} (i.e., activations and gradients generated by models from both sides) exchanges between the two sides. However, as LEO satellite networks offer only intermittent connectivity and low transmission rates between satellites and GS, conventional SL may fail to train models properly.
We set up motivating experiments, {as shown in Figure~\ref{subfig:motivating_setup},} via Starlink, the real-world commercial LEO satellite communication system,  in order to illustrate the challenge of developing SL on it. We use the well-known network measurement tool, \textit{Iperf}~\cite{tirumala1999iperf}, to test and record uplink/downlink rates. To evaluate SL performance, we adopt one of the most representative SL frameworks, SFL~\cite{thapa2022splitfed}, and take the well-known neural network VGG-16~\cite{simonyan2014very}  as the global model. We select the spatial modulation recognition dataset GBSense~\cite{gbsense} as training dataset.
Unless otherwise specified, this experimental setup remains the same thoughout the whole Section~\ref{sec:background_motivation}.

\begin{figure}[t]
  \centering
  \subfloat[Experimental setup. \label{subfig:motivating_setup}]
  {
    \includegraphics[width=0.455\linewidth]{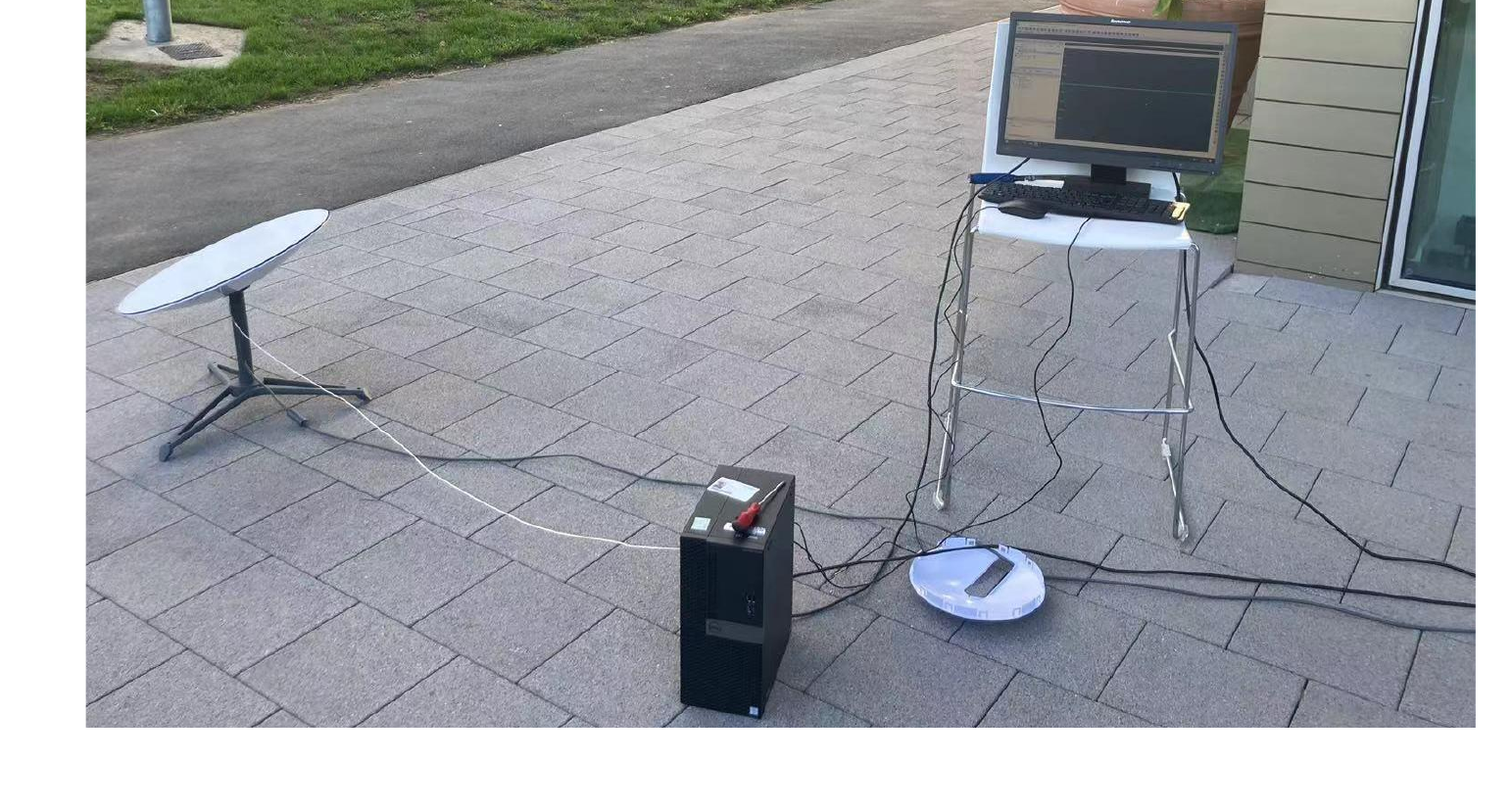}
  }
    \subfloat[Contact time. \label{subfig:idletime_ratio}]
  {
    \includegraphics[width=0.46\linewidth]{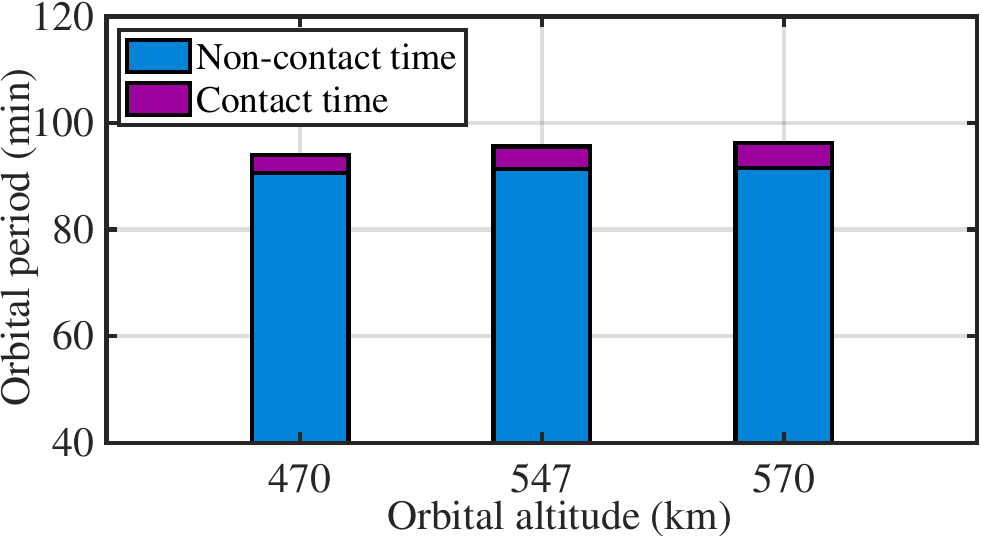}
  } \\
  \subfloat[CDF of uplink/downlink rates.
  \label{subfig:motivating_downlink}]
  {
    \includegraphics[width=0.477\linewidth]{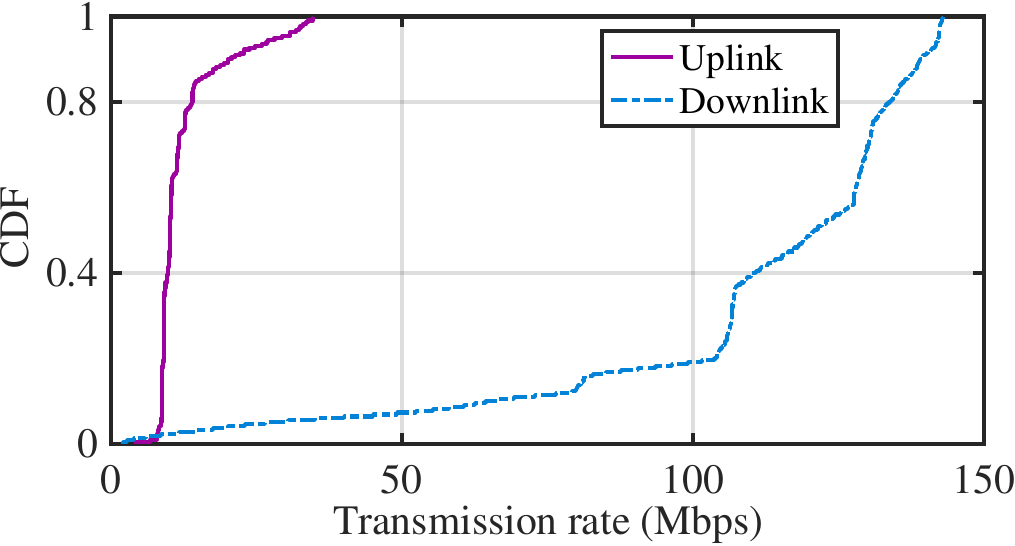}
  }
  \subfloat[Slow convergence speed. \label{subfig:idletime_impact}]
  {
    \includegraphics[width=0.477\linewidth]{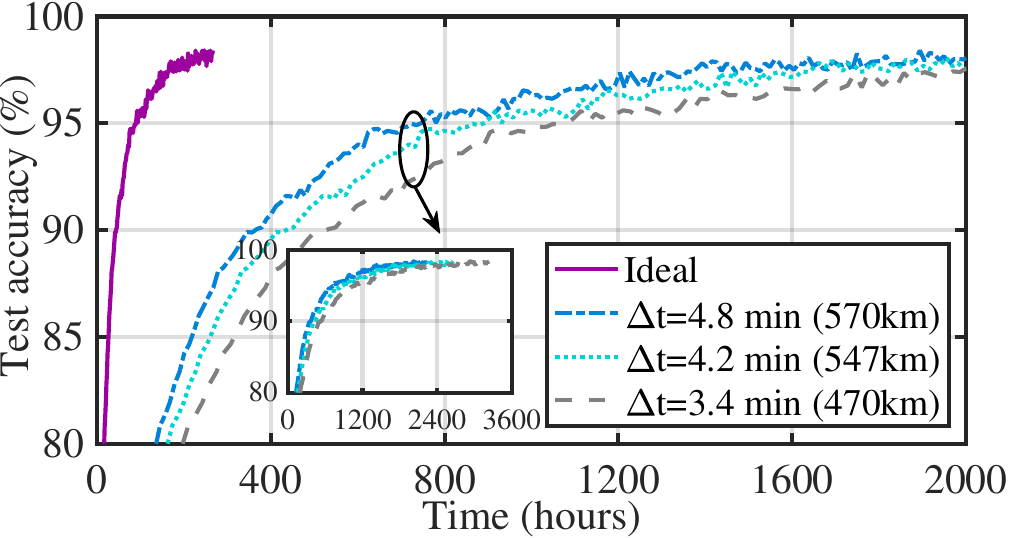}
  }
  \caption{Limited contact time (b) and transmission rates (c) between LEO satellites and GS become a major bottleneck for training SL models (d).}
  \label{fig:motivating_databalance}
\end{figure}

According to Starlink constellation configuration information ~\cite{starlink_gp},
the contact time between satellites and GS at different orbital altitudes can be derived. As shown in Fig.~\ref{subfig:idletime_ratio}, the contact time occupies only a small fraction (roughly 5\%) of its orbital period.  Meantime, we measure the transmission
rates of Starlink; the results shown
in Fig.~\ref{subfig:motivating_downlink} indicate
mean downlink and uplink rates of roughly 100 Mbps and 12 Mbps, respectively. Considering the sizes of smashed data being about 170\!~GB (equally shared between activation and gradient) when training VGG-16, a Starlink satellite at 547~\!km orbit with 4.2-minute contact time can only exchange approximately 29.5\% (resp. 3.5\%) of activations (resp. gradients) during the contact time with a GS.
Consequently, conventional SL cannot train VGG-16 completely.
Although specialized GS~(with multi-million US dollars cost~\cite{Devaraj2019PlanetHS,telesat}) offers Gbps- and hundreds of Mbps-level rates respectively for downlink and uplink, multiple tenants need to share these rates,
resulting in very limited per-tenant rates and thus severely restricting their smashed data exchanges. Moreover, recent work~{\cite{vasisht2021l2d2}} also utilizes distributed GSs of low-cost commodity hardware to receive satellite downlink signals. Therefore, results produced by our commercial GS
are sufficiently representative.



We then input the traces of Starlink~(e.g., contact time, uplink, and downlink rates) collected by us to our LEO satellite networks emulator (see Section~\ref{sec:implementation} for more details) to evaluate the performance of real SFL training in three different contact times. For the sake of comparison, we also consider an ideal case without contact time constraint, i.e., a satellite can continuously communicate with GS throughout any satellite orbital period.
The results in Figure~\ref{subfig:idletime_impact} demonstrate that the convergence speeds under intermittent connectivity are roughly ten-time slower than that of the ideal case.
This dramatic slowdown of model convergence occurs because models can be trained only during contact time.
Therefore, existing SL frameworks cannot be directly applied to LEO satellite networks, necessitating 
a LEO-tailored SL design.

\begin{figure}[t]
  \centering
  \subfloat[\# of sub-models.\label{subfig:performance_aggre_bias}]{
    \includegraphics[width=0.466\linewidth]{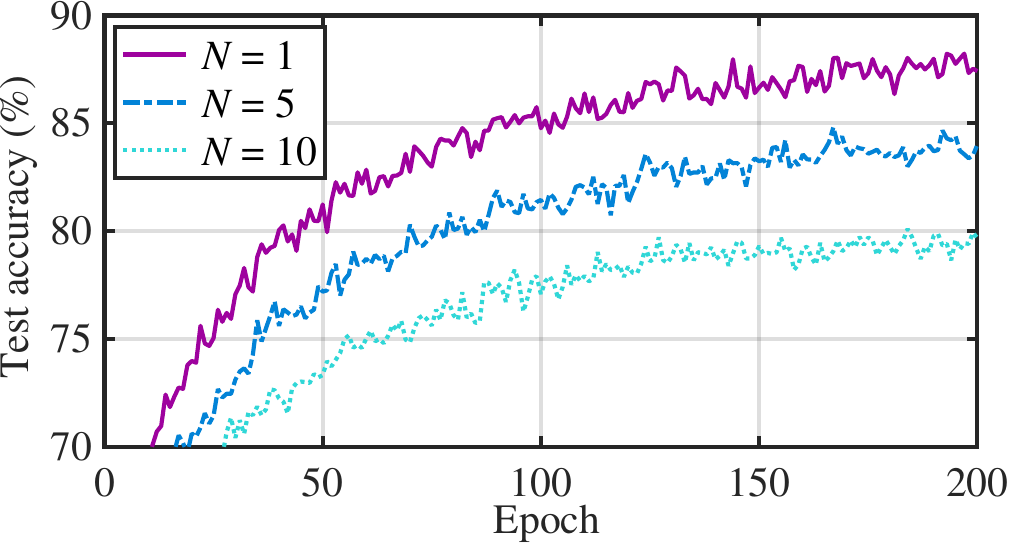}
  }
  \subfloat[Different labeling rates.\label{subfig:training_perfor_labeling}]
  {
    \includegraphics[width=0.466\linewidth]{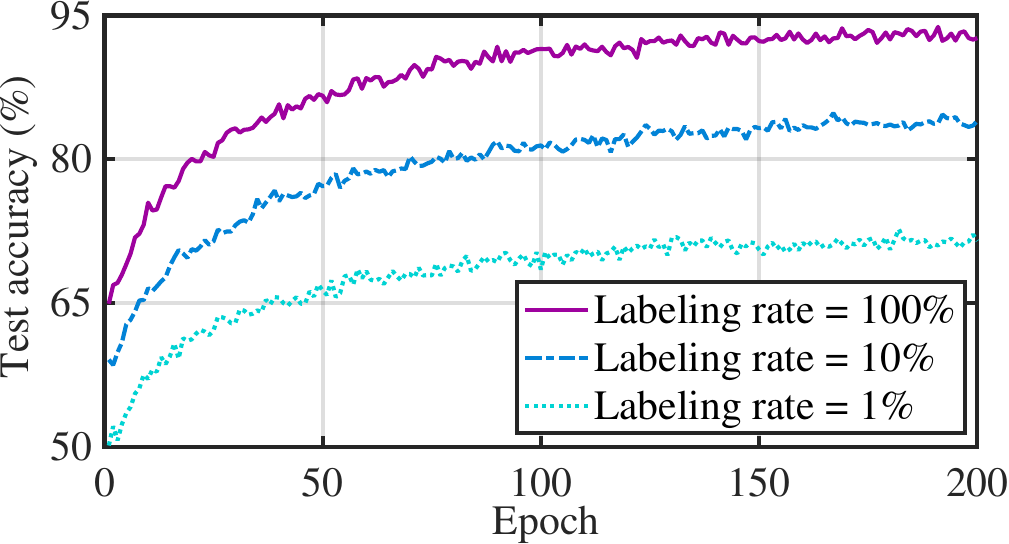}
  }
  \caption{The SL performance against (a) different number of sub-models involved in model aggregation and (b) distinct labeling rates.}
  \label{fig:motivating_labeling}
\end{figure}

\subsection{Accumulated Biases from SSL}\label{subsec:unlabeled_data}

Although SSL can lessen the need for labeled data in LEO satellite networks, it still introduces model bias due to incomplete training data. Unfortunately, the model aggregation process native to SL tends to accumulate and thus amplify such biases, further diminishing the model’s tolerance to them. 
To gain deeper insights into the impact of SSL on SL training performance, we build a naive semi-supervised SL using a direct combination of the seminal SSL $\pi$-model~\cite{laine2016temporal} and SFL. We then conduct motivating experiments with two different settings: i) different number $N$ of satellites~(i.e., $N = 1, 5, 10$) and a fixed labeling rate of 10\%; ii) three different labeling rates 1\%, 10\%, and 100\% for each satellite, and sub-models from 5 satellites are aggregated. 
%
We can clearly observe from Figure~\ref{subfig:performance_aggre_bias} that, with more satellite sub-models being aggregated,
the test accuracy decreases significantly: 
the global model aggregating 10 sub-models has 8.5\%, and 4.9\% accuracy losses compared to the counterparts with 1 and 5 sub-models being aggregated. 
As expected, the global model with complete labeling~(i.e., labeling rate = 100\%), significantly outperforms others~(sub-models with 1\% or 10\% labeling rate) in Figure~\ref{subfig:training_perfor_labeling}. 

The results reveal two potential reasons for the infeasibility of directly combining SL and SSL. On the one hand, the sub-models from different satellites contribute their own biases to the global model during aggregation. These individual biases mostly do not offset each other but rather get accumulated within the global model, leading to a significant decrease in the test accuracy (on unseen data) as the number of sub-models (resp. labeling rate) increases, indicating a degradation in the model generalization. On the other hand, the limited labeled data of each satellite forces its model gradients closely aligned to the local optimal direction, rather than the global one. Consequently, our preliminary tests seem to suggest a risk of drastic drop in performance upon aggregating a much larger amount of sub-models~(e.g., $N$ $\ge$ 100), indicating the possibility of overfitting on individual client-side datasets.

\begin{figure}[b]
  \centering
  \subfloat[Class imbalance.\label{subfig:motivating_classimb}]{
    \includegraphics[width=0.472\linewidth]{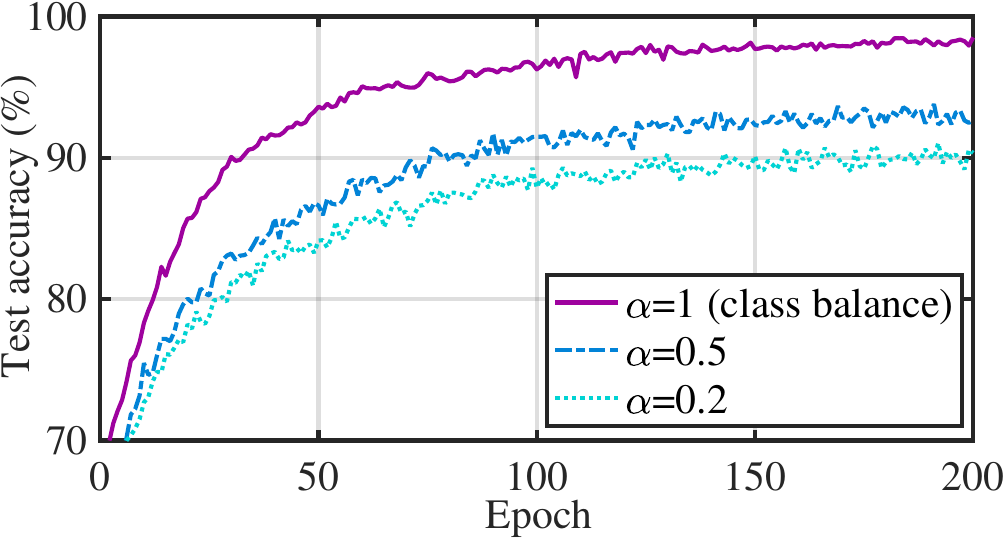}
  }
  \subfloat[Quantity imbalance.\label{subfig:motivating_quantimb}]
  {
    \includegraphics[width=0.472\linewidth]{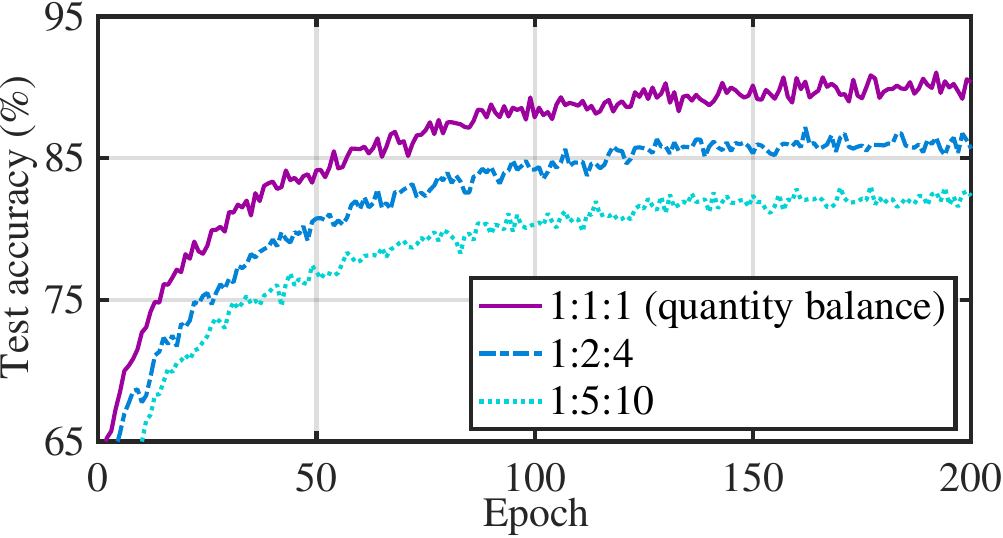}
  }
  \caption{The impact of data class imbalance (a) and quantity imbalance (b) on SL performance.}
  \label{fig:motivating_labeling}
\end{figure}

\begin{figure*}[t!]
\centering
\includegraphics[width=16.5cm]{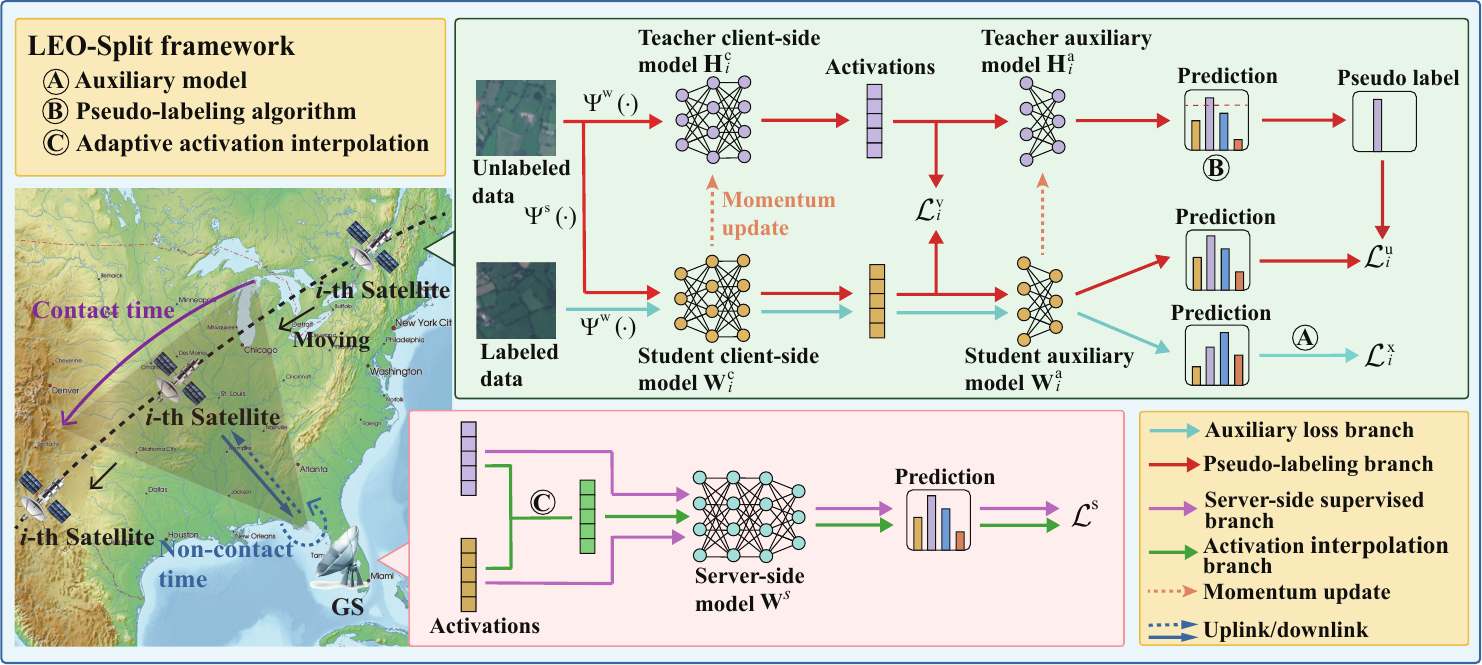}
\caption{An overview of \name architecture.}
\label{fig:s3l_framwork}
\end{figure*}

\subsection{Data Class and Quantity Imbalance}\label{subsec:data_imbalance}
Data collected by LEO satellites are inherently heterogeneous, mainly in terms of class imbalance and quantity imbalance~\cite{zhai2023fedleo}.
Satellites in different orbits may cover very distinct geographical regions, each having its unique characteristics (e.g., climate, geology, and vegetation types). Therefore, classes covered by individual satellite can be highly imbalance as several classes can go missing in the dataset collected by a certain satellite.
Additionally, disparities in storage and sensing capabilities limit the amount of data each satellite can retain, causing data quantity imbalance across satellites.
To understand how data heterogeneity impacts the performance of SL, we conduct two experiments
for class and quantity imbalance.
%
%
For class imbalance, we leverage Dirichlet data partitioning~\cite{chen2022towards} to set three levels of class imbalance controlled by parameter $\alpha$, where a larger $\alpha$ indicates a more balanced class across satellites.
For quantity imbalance, we configure three local dataset size ratios of  1:1:1 (balance), 1:2:4 (mild imbalance), and 1:5:10 (severe imbalance), while keeping $\alpha = 0.2$.

Figure~\ref{subfig:motivating_classimb} and~\ref{subfig:motivating_quantimb} both indicate a certain level testing accuracy degradation due to class imbalance (e.g., $\alpha = 0.2$ causes an accuracy drop of 8.1\%)
and quantity imbalance (an accuracy drop of 8.5\% under severe imbalance).
%
Though the negative impact of data heterogeneity may appear to be mild, it can still be significant for certain LEO-enabled applications~\cite{jardak2022potential,7060481}.
Since satellites train their sub-models on local datasets, the existence of class imbalance (e.g., skewing towards certain classes)
may result in overfitting the dominant classes while forgetting the under-represented ones~\cite{huang2022learn,pillutla2022federated,qu2022rethinking}.
Similarly, quantity imbalance can mislead the updated global model towards data-rich satellites while neglecting the data-poor ones~\cite{li2023revisiting,abay2020mitigating}; the model performance can hence be degraded if the neglected part actually contain crucial information.

\section{System Design}\label{sec:s3l_frame}

\subsection{Overview}\label{subsec:preli_overview}
%



In this section, we present \name, a semi-supervised split learning framework specifically tailored for LEO satellite networks. To combat the aforementioned challenges
faced by deploying \newrev{deep learning} for LEO satellite networks, we meticulously design three key components for \name:

\begin{itemize}
    \item  
    To significantly eliminate the dependence of SL training on continuous satellite-GS connectivity, we design an \textit{auxiliary model} (Section~\ref{sssec:aux}) with the same output dimension as the global model,
    allowing the satellite to train independently.
    %
    %
    %
    \item To combat the cumulative biases and the catastrophic forgetting induced by data heterogeneity, we propose a \textit{
    pseudo-labeling algorithm} (Section~\ref{subsec:pseudo_labelling}); it dynamically adjusts a threshold based on data distribution across satellites throughout model training, so as to 
    achieve class-balanced model training.  
    %
    \item {To effectively mitigate the performance degradation introduced by overfitting, we develop an \textit{adaptive activation interpolation} (Section~\ref{ssec:server_side})
    that interpolates the limited activations at GS to minimize the discrepancy in training data volume between satellites and GS.}

%
\end{itemize}

As shown in Figure~\ref{fig:s3l_framwork}, the global model $\bf W$ is split into client-side and server-side sub-models, ${\bf W}^{\mathrm{c}}_i$ and ${\bf W}^{\mathrm{s}}$, respectively, with ${\bf W}^{\mathrm{c}}_i$ deployed on the $i$-th satellite and ${\bf W}^{\mathrm{s}}$ on the GS. The training workflow of \name for one training round (defined as one satellite orbital period)
follows three steps: i) each satellite utilizes the auxiliary model, and pseudo-labeling algorithm to train its sub-model, ii) satellites transmit activations to the GS as much as possible during contact time, and finally iii) GS leverages the adaptive activation interpolation to achieve data argumentation and then train its sub-model for entering the next training round. This last step also involves aggregating sub-models into the global one, but it can be conducted less often.
Note that \name is built upon a common SL framework~\cite{thapa2022splitfed}, where aggregation refers to a weighted average of several sub-models, similar to the behavior of FedAvg~\cite{mcmahan2017communication} but only acting on part of the global model that gets trained on the client side.

\subsection{Client-side Design} \label{ssec:client-side}

%
%
%
We hereby set up the nomenclature for the client-side. To train on partially labeled data, the client-side of \name leverages a classical SSL framework, Mean Teacher~\cite{tarvainen2017mean}, consisting of two branches:
i) \textit{student} model ${{\bf{\widetilde W}}}^{\mathrm{c}}_{i} = [ {{\bf{W}}^{\mathrm{a}}_{i};{{\bf{W}}^{\mathrm{c}}_{i}}} ]$ (see Section~\ref{sssec:aux}), and ii) \textit{teacher} model ${\bf{\widetilde H}}^{\mathrm{c}}_{i} = [ {{{\bf{H}}^{\mathrm{a}}_{i}};{{\bf{H}}^{\mathrm{c}}_{i}}} ]$ (see Section~\ref{subsec:pseudo_labelling}), where ${\bf{W}}^{\mathrm{c}}_{i}$ (resp. ${{\bf{H}}^{\mathrm{c}}_{i}}$) and ${\bf{W}}^{\mathrm{a}}_{i}$ (resp. ${{\bf{H}}^{\mathrm{a}}_{i}}$) are student~(resp. teacher) sub-model and \textit{auxiliary} model, respectively; the auxiliary model is meant to cope with the intermittent connectivity. 
%
%
%
%
For each training round, the $i$-th satellite aims to minimize its loss function
${{\mathcal{L}}^{\mathrm{c}}_{i}} = {{\mathcal{L}}^{\mathrm{x}}_{i}} + {\lambda ^{\mathrm{u}}}{{\mathcal{L}}^{\mathrm{u}}_{i}} + {\lambda ^{\mathrm{v}}}{{\mathcal{L}}^{\mathrm{v}}_{i}}
$,
where ${{\mathcal{L}}^{\mathrm{x}}_{i}}$, ${{\mathcal{L}}^{\mathrm{u}}_{i}}$, and ${{\mathcal{L}}^{\mathrm{v}}_{i}}$ represent the auxiliary loss, unsupervised loss, and contrastive loss, respectively, with details provided in the following sections. The hyper-parameters $\lambda^{\mathrm{u}}$ and $\lambda^{\mathrm{v}}$ are balance weights for ${{\mathcal{L}}^{\mathrm{u}}_{i}}$ and ${\mathcal{L}}^{\mathrm{v}}_{i}$. During training, 
the student model updates its weights by minimizing $\mathcal{L}^{\mathrm{c}}_{i}$, while the teacher model's weights are updated using the exponential moving average (EMA)~\cite{tarvainen2017mean} of the student model's weights.
For the next training round, the weights of the updated teacher model are assigned to the student model as the initial weights.

\subsubsection{{Auxiliary Model}}\label{sssec:aux}
%
%
As explained in Section~\ref{subsec:limit_contact}, limited contact time and transmission rate cause training failure during satellite-GS non-contact time. To overcome this challenge, we empower satellites with independent client-side training capability during non-contact time. As shown in Figure~\ref{fig:auxiliary_model}, for conventional SL, the output layer is positioned at the final layer of the model, necessitating data flow to go through the entire model, thus incurring frequent data exchanges between clients and the server. 
To eliminate the SL's inherent reliance on continuous data exchange, we borrow the idea of multi-exit neural networks~\cite{10229092} to customize early outputs in the shallow layers for facilitating flexible partial model updates. To this end, we add two layers (one CNN and one FC) before the output layer, and we align their input dimensions with the client-side sub-model's last layer while matching their output dimensions with that of the server-side sub-model. This compact neural model, termed \textit{auxiliary} model, thus allows \name to be trained without fully interacting with the GS.

In order to properly train the auxiliary model, we construct an auxiliary loss against the student model $\widetilde{\bf W}_{i}^{\mathrm{c}}$ as:
%
\begin{align}\label{auxiliary_loss_function}
\small {\mathcal L}^{\mathrm{x}}_{i} = \frac{1}{| {\mathcal D}_i^{\mathrm{L}} |} \sum\limits_{k = 1}^{| {\mathcal D}_i^{\mathrm{L}} |} H(y_{i,k}^{\mathrm{L}},p({\Psi}^{\mathrm{w}}({\bf x}_{i,k}^{\mathrm{L}});{\bf \widetilde W}^{\mathrm{c}}_{i})) ,
\end{align}
where ${\mathcal{D}}^{\mathrm{L}}_i  = \{ {\bf{x}}^{\mathrm{L}}_{i,k},y^{\mathrm{L}}_{i,k} \}_{k = 1}^{{| {\mathcal D}_i^{\mathrm{L}} |}}$ is the local labeled dataset of the $i$-th satellite,
${\bf{x}}^{\mathrm{L}}_{i,k}$ and ${y}^{\mathrm{L}}_{i,k}$ denote the $k$-th input data and its corresponding label, ${\Psi}^{\mathrm{w}} (\cdot)$ is the weak data augmentation operation (e.g., using only flip-and-shift data augmentation~\cite{sohn2020fixmatch}), 
${p}( {x ; w} )$ maps the relationship between input data $x$ and its predicted class distribution given model parameter $w$, and $H(\cdot , \cdot)$ denotes the cross-entropy.} Subsequently, each satellite may independently trains its sub-model via minimizing  auxiliary loss ${{\mathcal L}^{\mathrm{x}}_{i}}$ 
during non-contact time.

\begin{figure}[t]
\centering
\includegraphics[width=.94\columnwidth]{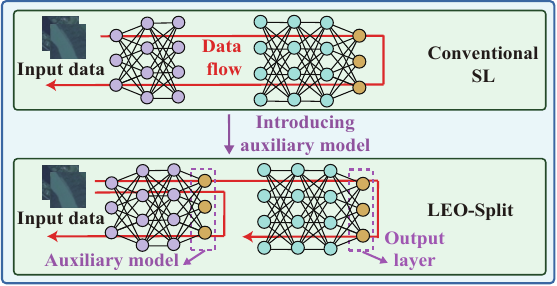}
\caption{
The comparison of sub-model training for conventional SL and \name. }
\label{fig:auxiliary_model}
\end{figure}

Recall that satellites can transmit only a small fraction (e.g., 10\%) activations to the GS during contact time, due to the limited downlink rate. To this end, \name randomly selects activations to saturate the downlink during the contact time, leaving the issue of incomplete interaction handle by the server side (see Section~\ref{ssec:server_side}).
Intuitively speaking, such incomplete interactions could severely slow down the convergence: at least 10 rounds (given 10\% downloadable activations every round) would be needed on average to make one epoch progress in training.
%
%
Fortunately, because the activations are inherently sparse, our later results (see Section~\ref{sec:simulation results}) demonstrate that \name can achieve much faster convergence than the expected linear scaling.
The rationale behind this promising outcome goes very close to how dropout~\cite{srivastava2014dropout} leverages a small subset of neurons in a model to achieve better performance. In other words, since deep neural models involve a huge amount of elements (e.g., neurons and activations), the chance of having only a small fraction of them being independent is high. Therefore, \name's partial yet random updates could go a much longer way than one would expect.

\subsubsection{{Pseudo-Labeling Algorithm}}\label{subsec:pseudo_labelling}
According to the discussions in Sections~\ref{subsec:unlabeled_data} and~\ref{subsec:data_imbalance}, directly combining SSL with SL (as in Section~\ref{sssec:aux}) can cause accumulated biases and catastrophic forgetting. Fortunately, our earlier discussions also reveal that the shortage of labeled data and data heterogeneity across satellites are two fundamental reasons: solving them could lead to a seamless welding of SL and SSL. Common strategies to handle these problems often resort to certain forms of label generation~\cite{sohn2020fixmatch}; generating meaningful labels apparently compensates labels shortages, while class imbalance can also be alleviated if classes without (or with few) labeled data can acquire sufficient labeling.

However, a single model cannot generate pseudo-labels by itself, otherwise using them for training will lead to zero yet meaningless loss.
To this end, a commonly adopted approach is to involve another model to generate pseudo-labeled data for training the original model, and the teacher-student co-training framework is arguably the most successful way to coordinate these two models~\cite{tarvainen2017mean}. In particular, the additional model and the original model are treated as \textit{teacher} and \textit{student} models, respectively. As shown in  Figure~\ref{fig:s3l_framwork}, unlabeled data is put into the teacher model to predict a class label with a confidence score that reflects the probability of that prediction being correct. The confidence score is then compared against a predefined \textit{threshold}: the label is chosen as the pseudo-label if the score exceeds the threshold; otherwise it is dropped. \newrev{These pseudo-labeled are taken by the student model to enhance its semi-supervised training.} 
%

In terms of defining the threshold, state-of-the-art proposals mostly stick to a static threshold that remains constant during the whole co-training process~\cite{sohn2020fixmatch,berthelot2019mixmatch}. This is apparently not reasonable in the face of inherent class imbalance,
because it fails to \newrev{adapt to discrepancy in data distribution across clients in distributed training and distribution variation during individual model training.} 
\begin{figure}[t]
  \centering
  \subfloat[Accumulated biases.\label{subfig:motivating_quantimb_1}]{
    \includegraphics[width=0.472\linewidth]{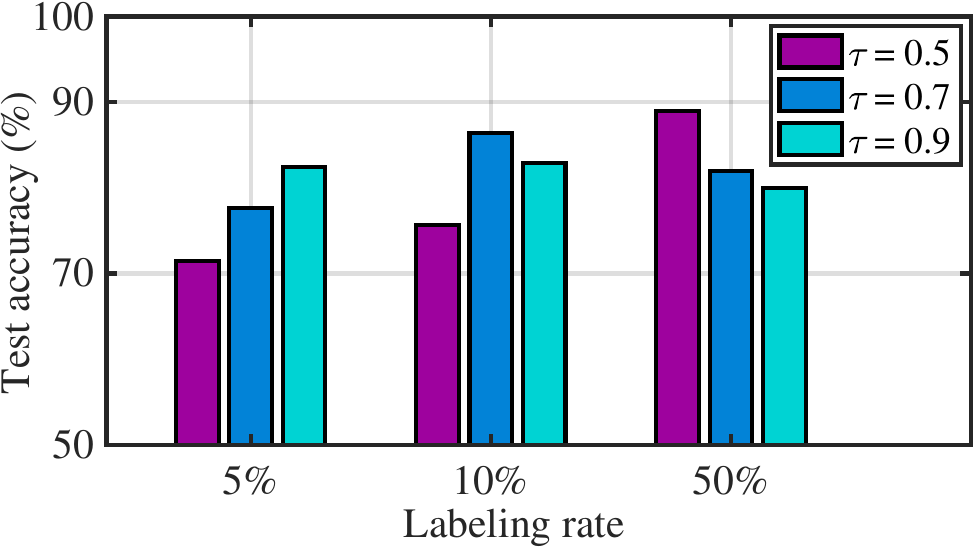}
  }
  \subfloat[Data heterogeneity.\label{subfig:motivating_classimb_1}]
  {
    \includegraphics[width=0.472\linewidth]{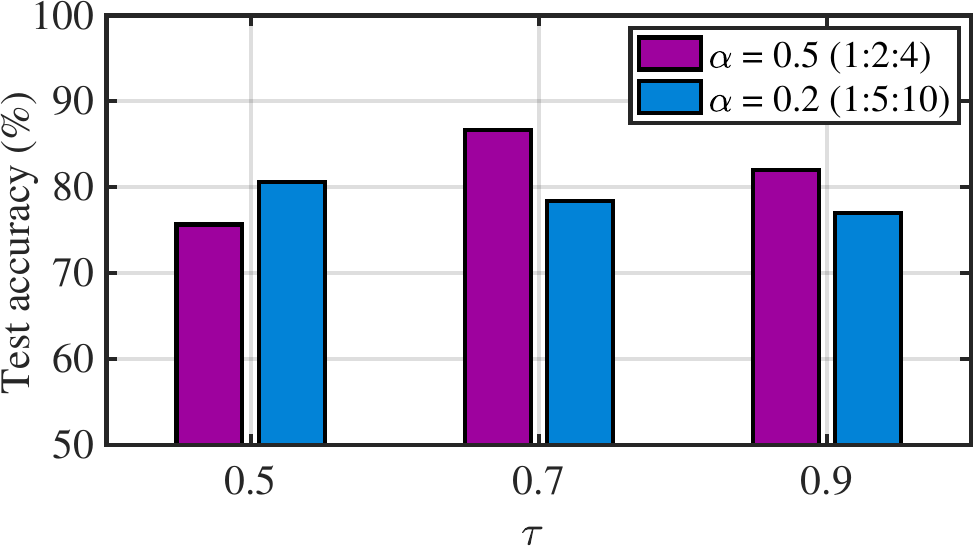}
  }
  \caption{Accumulated biases (a) and data heterogeneity (b) pose substantial barriers to the design of the pseudo-label threshold.}
  \label{fig:motivating_databalance}
\end{figure}
We conduct two experiments using a conventional fixed-threshold pseudo-labeling method~\cite{sohn2020fixmatch}, and set the fixed thresholds as 
$\tau(m)$ $\in$ $\{$$0.5$, $0.7$, $0.9\}$ 
for any class.
Figure~\ref{subfig:motivating_quantimb_1} illustrates 
the optimal thresholds vary under different labeling rates (i.e., 5\%, 10\%, and 50\%), and Figure~\ref{subfig:motivating_classimb_1} shows that different thresholds show varying performance in the two heterogeneous data settings ($\alpha = 0.2, 0.5$ with local dataset ratios 1:5:10 and 1:2:4, respectively). The results show that static thresholds \newrev{fail to perform adequately under the shortage of labeled data and data heterogeneity, implying the incompetence of static threshold to adapt to diverse} data distributions.  

\newrev{For enhancing adaptability, dynamically adjusting thresholds to suit different data distributions appears to be imperative.
Consequently, we focus on fine-tuning the threshold to balance the data heterogeneity that includes both class and quantity imbalance.
Intuitively, for class imbalance, we set lower thresholds for classes with more data, generating fewer pseudo-labels to counteract imbalance; for data quantity imbalance, we use higher thresholds for unlabeled datasets with more training data generating fewer pseudo-labels. Extending from these basic intuitions,}
we propose a pseudo-labeling algorithm to achieve the adaptive threshold $\tau_{i,t}(m)$ for the $m$-th class of the $i$-th satellite at the $t$-th round. \textbf{Algorithm~\ref{alg:dataset_status}} consists of \textsf{local dataset status collection}, and \textsf{adaptive threshold customization} stages. In the first stage, the $i$-th satellite uses the teacher model to generate pseudo-labeled data, and calculates the number of labeled and pseudo-labeled data $\theta^{\text{L}}_{i,t}(m)$ + $\theta^{\text{U}}_{i,t}(m)$ for the $m$-th class (\textsf{Lines~4} to \textsf{5}). 
Afterward, all satellites transmit their $\theta_{i,t}(m)$ to GS during contact time. In the second stage, we calculate the empirical class distribution of the $m$-th data class,
and the proportion of its amount within the total amount (\textsf{Lines~8} to \textsf{9}).
%
Meanwhile, we introduce the standard deviation of data classes to adjust our thresholds, speeding up the generation of balanced pseudo-labels in cases of extreme imbalance (\textsf{Line~10}). 
The $\tau^h$ ensures that $\tau _{i,t}(m)$ does not become extremely high, avoiding the loss of pseudo-labeling capability for certain classes (\textsf{Lines~11} to \textsf{12}). 

\RestyleAlgo{ruled}
\LinesNumbered
\begin{algorithm}[t]
\caption{{Pseudo-Labeling Algorithm.}}\label{alg:dataset_status}
\setstretch{1}
\small
\SetKwInOut{Input}{Require}
\SetKwFunction{Fns}{Local Dataset Status Collection}
\SetKwFunction{Fg}{Adaptive Threshold Customization}
\SetKwProg{Fn}{}{:}{}
\SetKwInOut{Output}{Output}

\Input{
The upper bound of threshold ${\tau ^{\mathrm{h}}}$, and pre-defined base threshold $\tau$.}
\KwData{The labeled and unlabeled datasets of $i$-th satellite ${\mathcal{D}}^{\mathrm{L}}_i  = \{ {\bf{x}}^{\mathrm{L}}_{i,k},y^{\mathrm{L}}_{i,k} \}_{k = 1}^{{| {\mathcal D}_i^{\mathrm{L}} |}}$ and ${\mathcal{D}}^{\mathrm{U}}_i  = \{ {{{\bf{x}}^{\mathrm{U}}_{i,k}}} \}_{k = 1}^{{{| {{\mathcal D}_i^{\mathrm{U}}} |}}}$ .}
\Output{The customized threshold for each class $m$ in the training round $t$, ${\tau_{i,t}}(m)$.}

\For{training round $t = 1,2,\ldots, T $}{
\Fn{\Fns}{
    \For{the $i$-th satellite in parallel}{
    $\theta_{i,t}^{\mathrm{L}}(m)$ $\leftarrow$ \textsf{Count\_labeled\_data}$(m)$\;
    $\theta_{i,t}^{\mathrm{U}}(m)$ $\leftarrow$ \textsf{Count\_unlabeled\_data}$(m)$\;
    }
}

\Fn{\Fg}{
    $\theta_{i,t}(m)$ $\leftarrow$ $\theta_{i,t}^{\mathrm{L}}(m)$+$\theta_{i,t}^{\mathrm{U}}(m)$~~~// \textit{Count total data}\;
    ${\tilde q}_t(m) \leftarrow \frac{\sum\limits_{i = 1}^{N} \theta_{i,t}(m)}{{\sum\limits_{m = 1}^{M} {{\sum\limits_{i = 1}^N \theta_{i,t}(m)}} }}$ ~~// \textit{Empirical class distribution}\;
    ${\tilde r}_{i,t} \leftarrow \frac{{{\sum\limits_{m = 1}^{M} \theta_{i,t}(m)}}}{{\sum\limits_{i = 1}^N {{\sum\limits_{m = 1}^{M} \theta_{i,t}(m)}} }}$ \quad\quad\quad// \textit{Ratio of dataset size}\;
    $\tau _{i,t}(m)\! \leftarrow \!{\tilde r_{i,t} ( {{{\tilde q}_t}(m) + \tau  - \textsf{STD}_m ({{\tilde q}_t}(m))} )}$\;
    \quad\quad\quad\quad\quad\quad\quad\quad\quad\quad\quad\quad// \textit{Tailor adaptive threshold}\;
    \If{${\tau _{i,t}(m) > {\tau ^{\mathrm{h}}}}$}{
      $\tau _{i,t}(m) \leftarrow {\tau ^{\mathrm{h}}}$ \quad\quad\quad\quad// \textit{Threshold correction}\;
      }
}
}

\end{algorithm}

After generating the pseudo-labels, we can utilize these pseudo-labels to train the student model in a supervised manner. Therefore, we employ unsupervised loss ${{\mathcal{L}}^{\mathrm{u}}_{i}}$ with the same form as auxiliary loss ${{\mathcal{L}}^{\mathrm{x}}_{i}}$, to update the student model along with ${{\mathcal{L}}^{\mathrm{x}}_{i}}$.
Moreover, although the intrinsic features of low-confidence data are not obvious and thus cannot directly generate pseudo-labels, they can still facilitate model training by self-supervised training~\cite{oord2018representation}. The classical contrastive learning framework is used to explore their intrinsic features~\cite{oord2018representation} at the $i$-th satellite.  Low-confidence dataset $\overline {\mathcal D}_i^{\mathrm{U}}$ are inputted into student sub-model ${\bf W}^{\mathrm{c}}_{i}$ and teacher sub-model ${\bf H}^{\mathrm{c}}_{i}$ to extract their corresponding activations (a.k.a., features)  ${\bf Z}^{\mathrm{U}}_{i,k}$ and ${ \widetilde {\bf Z}}^{\mathrm{U}}_{i,k}$, respectively. The principle of our contrastive learning is to make the ${\bf Z}^{\mathrm{U}}_{i,k}$ and ${ \widetilde {\bf Z}}^{\mathrm{U}}_{i,k}$ of the same data as close as possible, and them from different data as far apart as possible.
Therefore, the self-supervised contrastive loss InfoNCE~\cite{oord2018representation} is in the following:
\begin{align}\label{contrastive_loss}
{ \mathcal L}^{\mathrm{v}}_{i} \!= \! - \!\! \sum\limits_{k = 1}^{| {\overline {\mathcal D}_i^{\mathrm{U}}} |}\! {\log } \frac{{\exp ({\bf{Z}}_{i,k}^{\mathrm{U}} \!\otimes\! \widetilde {\bf{Z}}_{i,k}^{\mathrm{U}}/\phi )}}{{\exp ({\bf{Z}}_{i,k}^{\mathrm{U}}\! \otimes\! \widetilde {\bf{Z}}_{i,k}^{\mathrm{U}}/\phi ) \!+\!\!\! \sum\limits_{j = 1}^{| \overline {\mathcal D}_i^{\mathrm{U}} |} {{\!\!{\mathbbm 1}_{(j \ne k)}}} {\rm{exp}}({\bf{Z}}_{i,k}^{\mathrm{U}} \!\otimes\! {\bf{Z}}_{i,j}^{\mathrm{U}}/\phi )}},
\end{align}
where $\otimes$ denotes dot product operation and $\phi$ is a temperature hyperparameter~\cite{liu2023adaptive}.

\subsection{Server-side Design} \label{ssec:server_side}

The client-side of \name comprises solely server-side model ${\bf W}^{\mathrm{s}}$. For each training round, GS updates the server-side model by minimizing the server-side loss ${{\mathcal L}^{\mathrm{s}}}$, with details provided in the following sections.
\begin{figure}[t]
  \centering
  \subfloat[Limited activations.\label{subfig:motivating_actnums}]{
    \includegraphics[width=0.465\linewidth]{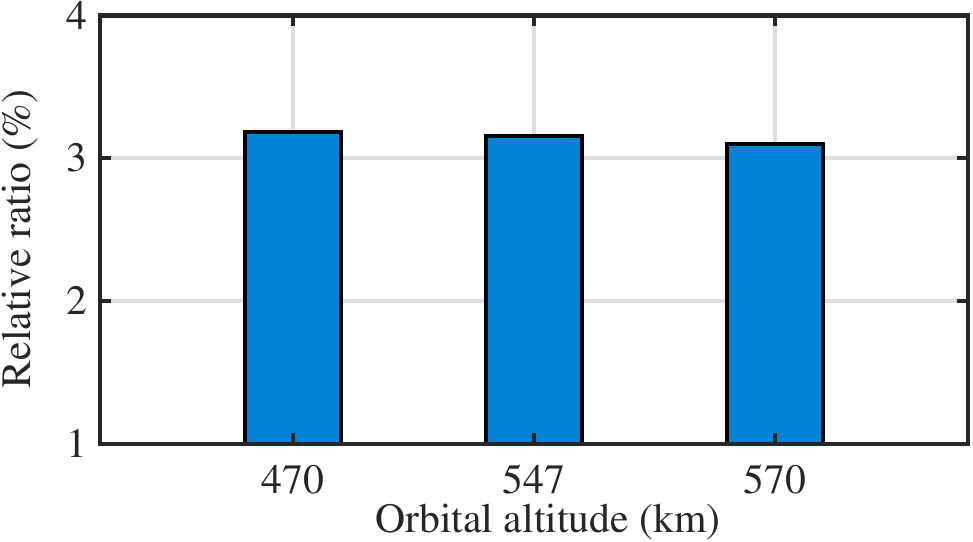}
  }
  \subfloat[Server-side overfitting.\label{subfig:overfitting_sub}]
  {
    \includegraphics[width=0.478\linewidth]{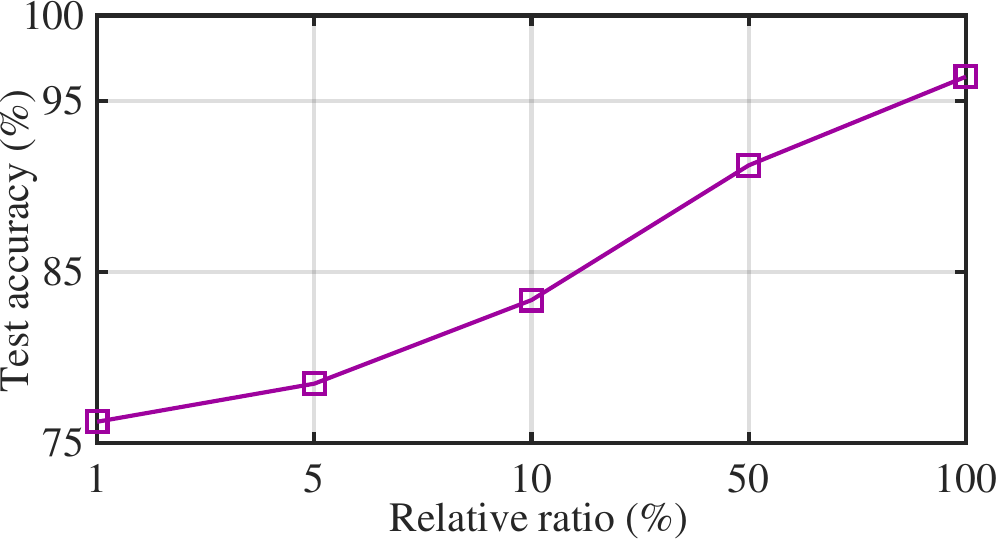}
  }
  \caption{The relative ratio across varying satellite orbital altitudes (a) and performance for test accuracy versus relative ratios (b).}
  \label{fig:overfitting_overall}
\end{figure}
Recalling the discussion in Section~\ref{sssec:aux}, auxiliary model empowers independent client-side sub-model updates to eliminate the training failure of non-contact time. However, the incomplete interaction between satellites and GS restricts the number of activations received by GS. As a result, the server-side model is forced to train on a very limited set of activations, \newrev{potentially} leading to 
server-side overfitting. Furthermore, due to the sparse nature of the activations, the class distribution represented by these activations may diverge significantly from the data distribution on the satellite. This discrepancy disrupts the consistency of sub-model training between satellites and the GS, further degrading overall training performance.
\rev{To investigate the impact of limited activations on SL performance, we conduct an experiment using auxiliary loss Eqn.~\eqref{auxiliary_loss_function}, where the relative ratio is the ratio of activations received by GS to the training data for client-side updates. Figure~\ref{subfig:motivating_actnums} shows that the relative ratios of satellites across diverse orbital altitudes are around 3\%, and Figure~\ref{subfig:overfitting_sub} illustrates that the scarcity of activations has a severe detrimental effect on SL performance, with a 3\% relative ratio causing an approximately 20\% drop in test accuracy.}

Though data interpolation~\cite{berthelot2019mixmatch} can be applied to address the insufficiency of training data, it is highly non-trivial to interpolate the activations caused by incomplete interaction, as the supplemented activations should help address the overfitting and data distribution discrepancies. One one hand, direly taking all acquired activations into interpolation may not be sensible, since 
only representative and independent activations may contribute to qualified reconstruction.
On the other hand, the data interpolation process should ensure the distribution of the interpolated activations closely aligned with that of client-side training data, thus avoiding the deterioration of the training performance caused by the serious deviation of the distribution. 
Therefore, we propose an adaptive activation interpolation scheme in \textbf{Algorithm~\ref{alg:aai}}; it  incorporates class-cycling activation selection to perform an even selection of activation across all classes, and then interpolates activations to align their distribution with that of the client-side data.

\vspace{.5em}
\textit{a) Class-cycling Activation Selection.} 
According to~\cite{goodfellow2016deep,yosinski2015understanding}, larger activations contribute more significantly to model updates, and thus, GS should prioritize downloading larger activations from satellites.  Additionally, to ensure data diversity of activations, selected activations from each class should be as evenly as possible. To achieve this, we adopt a class-cycling activation selection approach, where GS cycles through all classes and downloads the largest activation from each class in turn (\textsf{Lines~3} to \textsf{7}). It is noted that GS only downloads activations from more informative labeled and pseudo-labeled data.




\RestyleAlgo{ruled}
\LinesNumbered
\begin{algorithm}[t]
\caption{Adaptive Activation Interpolation.}\label{alg:aai}
\setstretch{1}
\small
\SetKwInOut{Input}{Require}
\SetKwFunction{Fns}{Class-cycling  Activation Selection}
\SetKwProg{Fn}{}{:}{}
\SetKwFunction{Fu}{Activation Interpolation}
\SetKwInOut{Output}{Output}

\Input{The total number of interpolations $J$, and parameter of the Beta distribution $\beta$.}
\KwData{The global dataset ${\mathcal D}^{\mathrm{A}} = \{ ({\bf{a}}_{k},y_{k})\} _{k = 1}^{|{\mathcal D}^{\mathrm{A}}|}$, where ${\bf{a}}_{k}$ and $y_{k}$ represent the $k$-th activations and its corresponding labels.}
\Output{Interpolated activation set ${{\mathcal D}_{(J)}^{\mathrm{a}}}$.}


\Fn{\Fns}{
    Initialize ${{\mathcal D}^{\mathrm{a}}} = \emptyset$\;
    \While{$r <$ $\mathrm{activation\_num}$}{
    \For{each class $m = 1,2,\ldots,M$}{
    $({\bf{a}}^{\prime}_{k},y^{\prime}_{k}) \!\!\leftarrow$ \textsf{Largest\_activation\_in\_class}(${{\mathcal D}^{\mathrm{A}}}, m$)\;    
    ${{\mathcal D}^{\mathrm{A}}} \leftarrow {{\mathcal D}^{\mathrm{A}}} \backslash \{ {\bf{a}}^{\prime}_{k},y^{\prime}_{k} \}$, ${{\mathcal D}^{\mathrm{a}}} \leftarrow {{\mathcal D}^{\mathrm{a}}} \cup \{ {\bf{a}}^{\prime}_{k},y^{\prime}_{k} \}$\;
    $r = r + 1$\;
    }
    }
}
\Fn{\Fu}{
    ${\mathcal D}^{\mathrm{a}}_{(0)} \leftarrow {\mathcal D}^{\mathrm{a}}$\;
    \For{interpolation round $j = 0,1,\ldots,J-1 $}{
    $({{\bf{a}}_{k_1, (j)}},{y_{k_1, (j)}})$ $\leftarrow$ \textsf{Random\_select}$({\mathcal D}^{\mathrm{a}}_{(j)})$\;

    \For{each class $m = 1,2,\ldots,M$}{
    ${\gamma ^{\mathrm{a}}_{(j)}}({m})$ $\leftarrow$ \textsf{Count\_data}$(m)$\;
    }
    $\alpha$ $\leftarrow$ \textsf{Beta\_distribution}$(\beta)$\;
    ${\hat \gamma ^{\mathrm{a}}_{(j)}}({{y_{k_1, (j)}}}) \leftarrow {\gamma ^{\mathrm{a}}_{(j)}}({{y_{k_1, (j)}}}) + \alpha$\;
    
    \For{$({{\bf{a}}_{k_2, (j)}},{y_{k_2, (j)}}) \in {{{\mathcal D}^{\mathrm{a}}_{(j)}}\backslash \{ {({{\bf{a}}_{k_1, (j)}},{y_{k_1, (j)}})} \}}$}{

    ${\hat \gamma ^{\mathrm{a}}_{(j)}}({{y_{k_2, (j)}}}) \leftarrow {\gamma ^{\mathrm{a}}_{(j)}}({{y_{k_2, (j)}}}) + 1 - \alpha$\;

    \For{each class $m = 1,2,\ldots,M$}{
    ${{\tilde q}_{(j)}^{\mathrm{a}}}( m; k_2 ) = \frac{{{{\hat \gamma_{(j)}^{\mathrm{a}}}(m)}}}{{{\sum\limits_{m = 1}^{\mathrm{M}} {{\hat \gamma_{(j)}^{\mathrm{a}}}(m)} }}}$~~~~~~// \textit{Class distribution}\;
    }
    }
    $({\bf{a}}^{*}_{k_2, (j)},y^{*}_{k_2, (j)}) \!\leftarrow\! \mathop {\arg \min }\limits_{k_2} {\textsf{MSE}_m  {( {\tilde q^{\mathrm{a}}_{(j + 1)}\!(m; k_2\!) , \tilde q(m)} )}}$\;
    $({\bar{\bf a}}_{(j)}, \!{\bar y}_{(j)}\!) \!\leftarrow\! \alpha ({{\bf{a}}_{k_1,(j)}}, {\!y_{k_1, (j)}}\!) \!+\! (1 \!- \!\alpha ) ({{\bf{a}}^{*}_{k_2, (j)}}, {y^{*}_{k_2, (j)}})$\;
    ${{\mathcal D}_{(j)}^{\mathrm{a}}} \leftarrow {{\mathcal D}_{(j-1)}^{\mathrm{a}}} \cup \{ {{\bar{\bf a}}_{(j)}}, {\bar y}_{(j)} \}$\;
    }
}

\end{algorithm}

\textit{b) Activation Interpolation.} To ensure consistent convergence between client-side and server-side sub-models, activation interpolation aims to increase the number of activations while retaining alignment between the data distributions of satellites and GS throughout the training process. Without loss of generality, we consider the $j$-th data interpolation operation at $t$-th training round for analysis. For notational simplicity, the training round index $t$ is omitted. 
The GS first randomly selects an activation and its label $({{\bf{a}}_{k_1, (j)}},{y_{k_1, (j)}})$  from the selected activation set ${\mathcal D}^{\mathrm{a}}_{(j)} = \{ ({\bf{a}}_{k, (j)},y_{k, (j)})\} _{k = 1}^{|{\mathcal D}^{\mathrm{a}}_{(j)}|}$, where ${\bf{a}}_{k, (j)}$ and $y_{k, (j)}$ represent the $k$-th selected activations and its corresponding label (\textsf{{Line~11}}). Then, GS computes the class empirical distribution ${{\tilde q}_{(j+1)}^{\mathrm{a}}}( m; k_1, k_2 )$ resulting from interpolating ${{\bf{a}}_{k_1, (j)}}$ with all remaining activations ${{\bf{a}}_{k_2, (j)}}$ in ${{{\mathcal D}^{\mathrm{a}}_{(j)}}}$ (\textsf{Lines~12} to \textsf{19}). Following this, ${{\bf{a}}^*_{k_2, (j)}}$ with the closest interpolated empirical class distribution to client-side class distribution $\tilde q(m)$ in Section~\ref{subsec:pseudo_labelling} (\textsf{Line~20}) is selected. 
Finally, ${{\bf{a}}_{k_1, (j)}}$ and ${{\bf{a}}^*_{k_2, (j)}}$, and their corresponding labels ${{y}_{k_1, (j)}}$ and ${{y}^*_{k_2, (j)}}$
are linearly interpolated and interpolated activation and label ${\bar{\bf{a}}}_{(j)}$ and ${\bar y}_{(j)}$ are updated to ${{{\mathcal D}^{\mathrm{a}}_{(j+1)}}}$ (\textsf{Lines~21} to \textsf{22}).

\begin{figure}[t]
\centering
\includegraphics[width=.95\columnwidth]{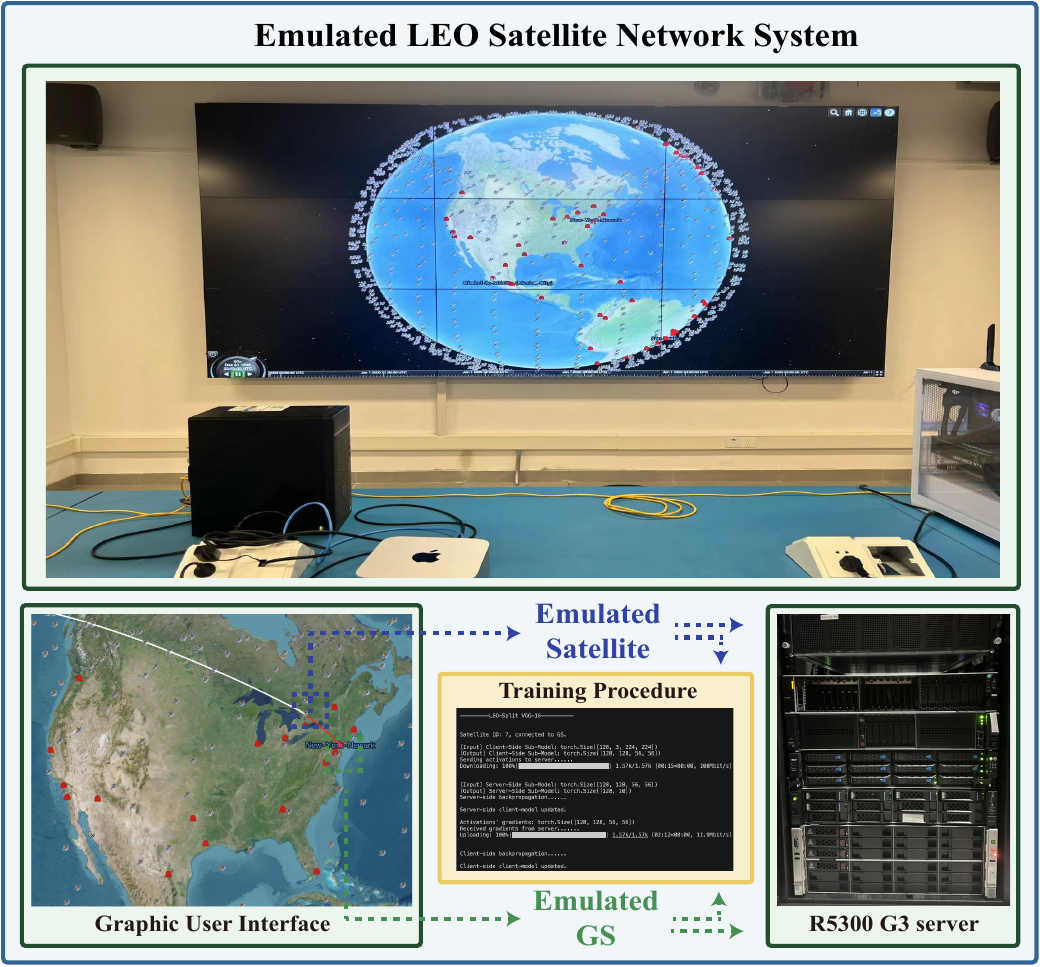}
\caption{
\name prototype and testbed. }
\label{fig:implementation}
\end{figure}

\section{{Implementation}}\label{sec:implementation}

In this section, we first elaborate on the implementation of \name, and we then introduce the experiment setup.

{
\subsection{Implementing \name}\label{sec:sys_implement}
We implement \name prototype for satellite-GS communication and neural network training based on a microservices architecture, as illustrated in Figure~\ref{fig:implementation}. The system is deployed in an H3C UniServer R5300 G3 server equipped with eight NVIDIA GeForce RTX 3090 GPUs, dual Intel Xeon Silver 4210R processors (10 cores, 2.84 GHz each), and 8×32 GB DDR4 RAM, running Ubuntu 18.04.6 LTS.
The software stack includes Python 3.7 and PyTorch 1.9.1, which are used for implementing the training of space modulation recognition and remote sensing image classification applications. 
We leverage a Kernel-based Virtual Machine~(KVM) to emulate a LEO satellite or a GS. The satellite and GS link is emulated and configured by \textit{Open vSwitch}~\cite{pfaff2015design}. The traffic conditions are controlled by \textit{tc}~\cite{beshay2015fidelity}, according to the traces of Starlink collected by us. 
The server-side and client-side sub-models are deployed within Linux containers (LXC) on the GSs, and satellites, respectively. 
The training procedures of both sub-models were executed on separate RTX 3090 GPUs within the R5300 G3 server. 
Real-world satellite orbits~(e.g., two-line element) are integrated into the system to enable real-time path computing and network routing, implementing dynamic connections between GS and satellites. This setup closely mirrors actual satellite communication links, and a 3D interface is used to visualize the connections and data exchanges between satellites and GSs via the emulation.

\subsection{Experiment Setup}\label{sec:experiment_setup}

\subsubsection{Dataset}\label{sec:simulation_setup_dataset}
We adopt the space modulation recognition dataset GBSense~\cite{gbsense} and the remote sensing image dataset EuroSAT~\cite{helber2018introducing} to evaluate the performance of \name.  
{The GBSense dataset consists of sampled signals with 13 modulation types such as BPSK, QPSK, and 8PSK, containing 16000 training and 4000 test samples. EuroSAT comprises 10 distinct categories of remote sensing images, including industrial, highway, and forest, with 21600 training samples and 5400 test samples.
We consider both IID (independent and identically distributed) and non-IID data settings in our experiment. In the IID setting, data samples are shuffled and evenly distributed to participating satellites. In the non-IID setting, we set $\alpha=0.5$ and divide satellites into three groups, with the local dataset size ratio of 1:2:4 across groups.


\subsubsection{Model}\label{sec:model_split_auxiliary}
To implement \name, we employ the well-known VGG-16 network~\cite{simonyan2014very} as the global model. VGG-16 is a classical deep convolutional neural network comprised of 13 convolution layers and 3 fully connected layers.  
In our experiment setup, the first four layers of VGG-16 are designated as the client-side sub-model, while the remaining layers are assigned as the server-side sub-model.

\subsubsection{Benchmarks}\label{sec:base_line}
To comprehensively evaluate the performance of \name, we compare \name against the following alternatives: 
\begin{itemize}
  \item \textbf{FM-SL} is the SL variant of FixMatch~\cite{sohn2020fixmatch}, which employs the pre-defined fixed threshold to generate high-confidence pseudo-labels on weakly augmented unlabeled data for guiding model training.
  \item \textbf{MM-SL} is the SL variant of MixMatch~\cite{berthelot2019mixmatch}, which utilizes distribution sharpening and pre-defined fixed threshold for generating pseudo-labels and enhances the raw data richness by interpolating labeled and unlabeled data. 
  \item \textbf{MT-SL} is the SL variant of Mean Teacher~\cite{tarvainen2017mean}, which updates the model based on the output consistency of the teacher and student model, and the teacher model is the exponential moving average of the student model.
  \item \textbf{$\pi$-SL} is the SL version of $\pi$-model~\cite{laine2016temporal}, which utilizes diverse data augmentation and dropout techniques to generate self-supervised signals, and employs Euclidean distance between outputs from different network branches for model training.
\end{itemize}

\subsubsection{Hyper-parameters}\label{sec:hyper-para}
%
In our experiments, we deploy a constellation of $N=10$ satellites orbiting the Earth} {and set labeling rates to 10\% by default unless specified otherwise.
The transmission rate between satellites and GS is consistent with Section~\ref{subsec:limit_contact}.} The computing capabilities of the satellites and GS are set to $1$ TFLOPS (peak performance of an iPhone 12 Pro~\cite{iphone}) and $4\times35.6$ TFLOPS ( peak performance of four NVIDIA RTX 3090), respectively. For the training process, we employ an SGD optimizer with a learning rate of 0.005 for each satellite, the batch size is set to 128, and $\tau^h$ is 0.95.

\section{{Performance Evaluation}}\label{sec:simulation results}



In this section, we evaluate the performance of \name from three aspects: i) comparisons with four benchmarks to demonstrate the superiority of \name; ii) investigating the impact of SL-related hyper-parameter on the model performance; iii) ablation study to show the necessity of each meticulously designed component in \name, including auxiliary model (AM), pseudo-Labeling algorithm (PA), and adaptive activation interpolation (AAI).

\subsection{Superiority of \name}


This section conducts a comprehensive comparison of \name against four benchmarks in terms of test accuracy and convergence speed. 

\begin{figure}[t]
  \centering
  \subfloat[GBSense (IID). \label{subfig:overall_curve_gbs_iid}]
  {
    \includegraphics[width=0.472\linewidth]{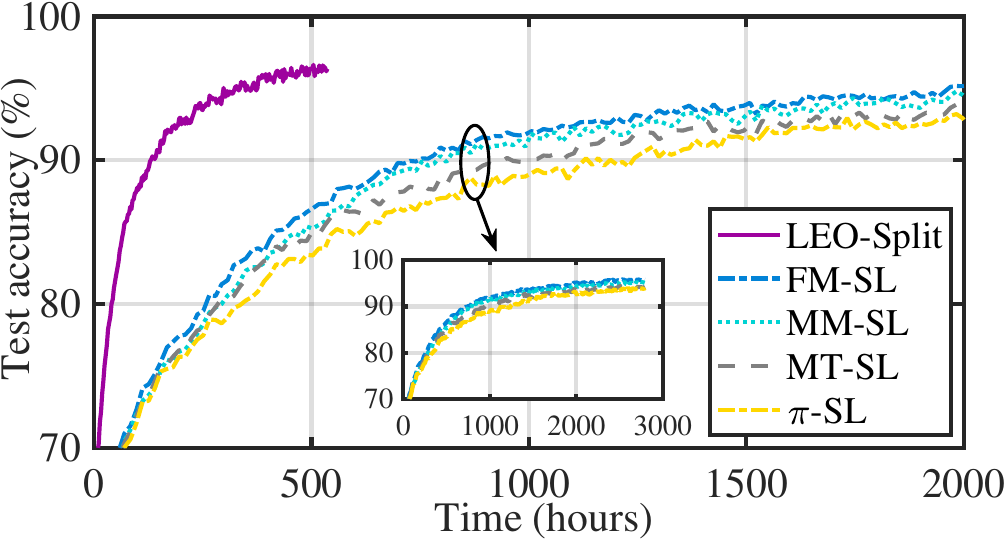}
  }
    \subfloat[GBSense (non-IID). \label{subfig:overall_curve_gbs_noniid}]
  {
    \includegraphics[width=0.472\linewidth]{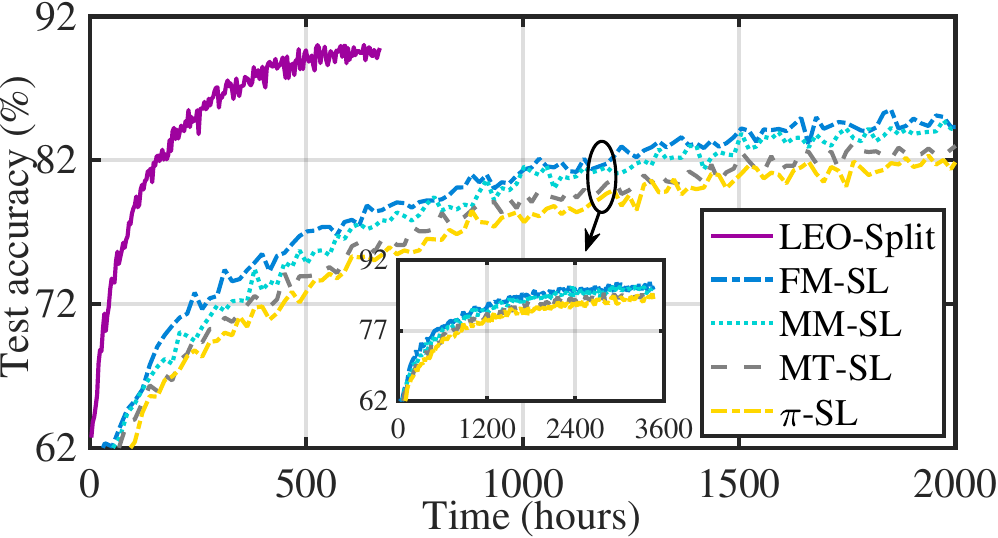}
  } \\
  \subfloat[EuroSAT (IID).
  \label{subfig:overall_curve_eus_iid}]
  {
    \includegraphics[width=0.472\linewidth]{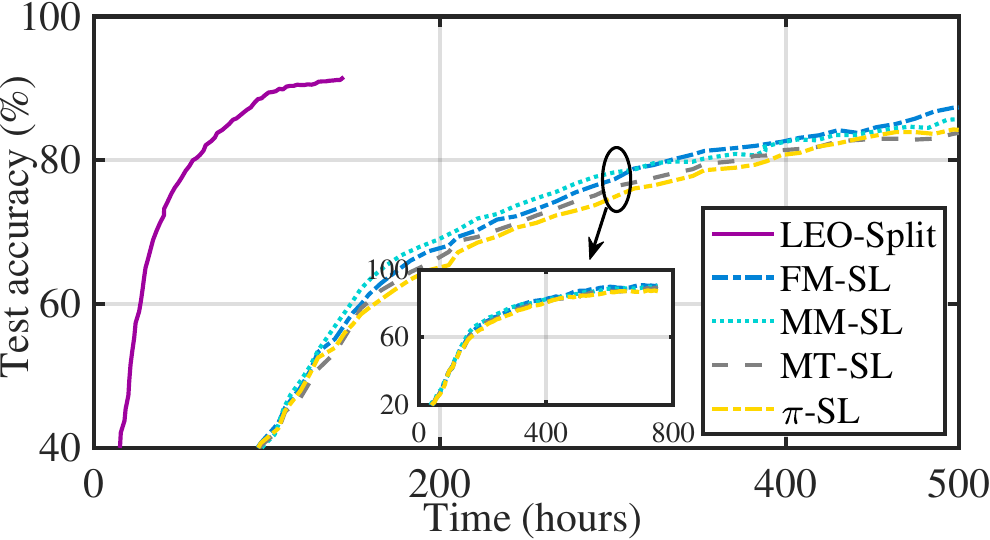}
  }
  \subfloat[EuroSAT (non-IID). \label{subfig:overal l_curve_eus_nonii}]
  {
    \includegraphics[width=0.472\linewidth]{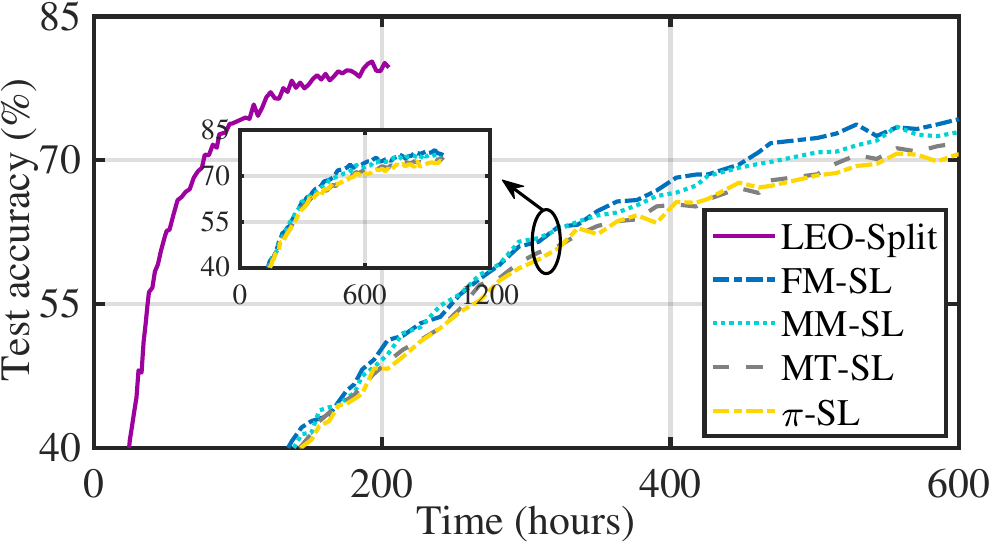}
  }
  \caption{Training performance on GBSense and EuroSAT under IID and non-IID settings of VGG-16.}
  \label{fig:simulation_overall_curve}
  \vspace{-2ex}
\end{figure}

\subsubsection{Training Performance of \name} {
Figure~\ref{fig:simulation_overall_curve} presents the superior performance of \name compared to four other benchmarks on GBSense and EuroSAT datasets. \name exhibits significantly faster convergence than the other benchmarks. {This is primarily attributed to the design of AM, which empowers independent client-side sub-model updates and thus overcomes the training failure.} In contrast, the
limited contact time and transmission rate
cause model training failure of four benchmarks in non-contact time
, thereby substantially slowing down model convergence. The sophisticated design of PA and AAI enables \name to achieve the highest converged accuracy. 
To be specific, {PA customizes pseudo-label thresholds for individual satellites to balance data class and quantity across satellites throughout model training}, and AAI linearly interpolates limited activations to increase the number of activations while guaranteeing data distribution consistency between satellites and the GS
to mitigate server-side overfitting. Conversely, MT-SL and $\pi$-SL demonstrate the worst training accuracy since they rely solely on feature-level consistency regularization and ignore the predictive probability distributions of models. Although FM-SL and MM-SL leverage the model's probability distributions for pseudo-labeling, their fixed pseudo-label thresholds fail to effectively combat data class and quantity imbalance, resulting in lower converged accuracy. 
}

\subsubsection{The Converged Accuracy and Time of \name}

Figure~\ref{fig:simulation_overall_bar} shows the converged accuracy and time of \name and four benchmarks on GBSense and EuroSAT datasets. It is seen that \name outperforms other benchmarks in both converged accuracy and time under IID and non-IID settings. For converged accuracy, Figure~\ref{subfig:overall_bar_gbsense_accu} and Figure~\ref{subfig:overall_bar_eus_accu} present that \name achieves {96.3\%} and {91.1\%} in test accuracy on the GBSense and EuroSAT datasets under the IID setting, nearly 1.6\% and 2.6\% higher on average than other benchmarks. This improvement is even more pronounced in the non-IID setting, approximately 4.8\% and 4.2\% on GBSense and EuroSAT datasets. {The reason for this is two-fold:} one is that PA is capable of effectively balancing data classes and quantity during pseudo-labeling to mitigate catastrophic forgetting and accumulated biases, and the other is that AAI expands the number of activations while guaranteeing data class distribution consistency between the satellites and GS. For converged time, Figure~\ref{subfig:overall_bar_gbsense_conv} and Figure~\ref{subfig:overall_bar_eus_conv} show that \name converges at a staggering 4.4 and 4.6 (4.6 and 4.7) on average times faster than other benchmarks on the GBSense (EuroSAT) dataset under IID and non-IID settings. {This significant improvement in convergence speed is attributed to} \name's AM design, which eliminates the dependence on continuous satellite-GS connectivity, thereby preventing training failure of the satellite-GS non-contact time.

\begin{figure}[t]
  \centering
  \subfloat[Test accuracy (GBSense). \label{subfig:overall_bar_gbsense_accu}]
  {
    \includegraphics[width=0.465\linewidth]{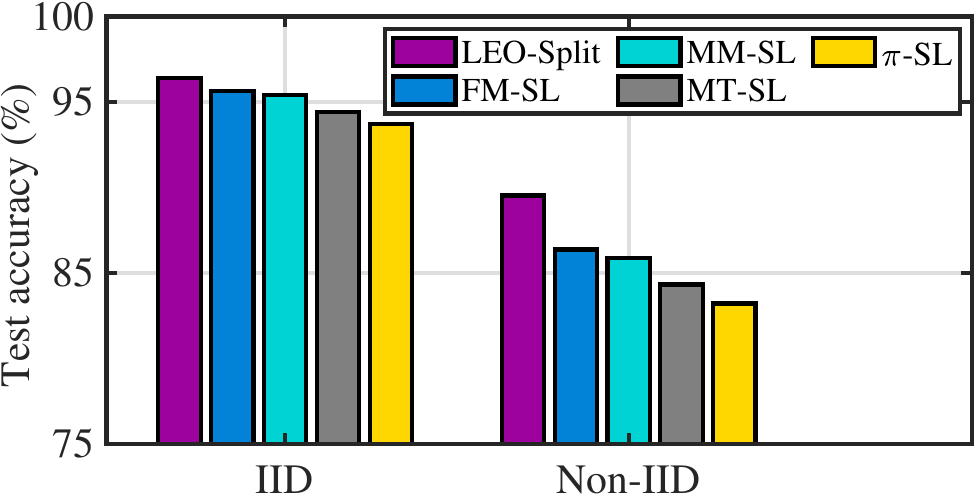}
  }
    \subfloat[Converged time (GBSense). \label{subfig:overall_bar_gbsense_conv}]
  {
    \includegraphics[width=0.476\linewidth]{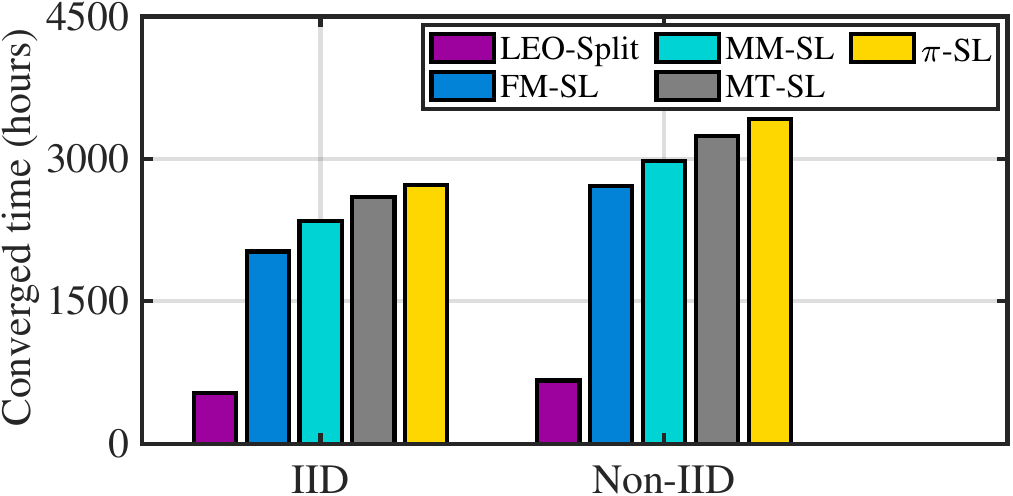}
  } \\
  \subfloat[Test accuracy (EuroSAT).
  \label{subfig:overall_bar_eus_accu}]
  {
    \includegraphics[width=0.465\linewidth]{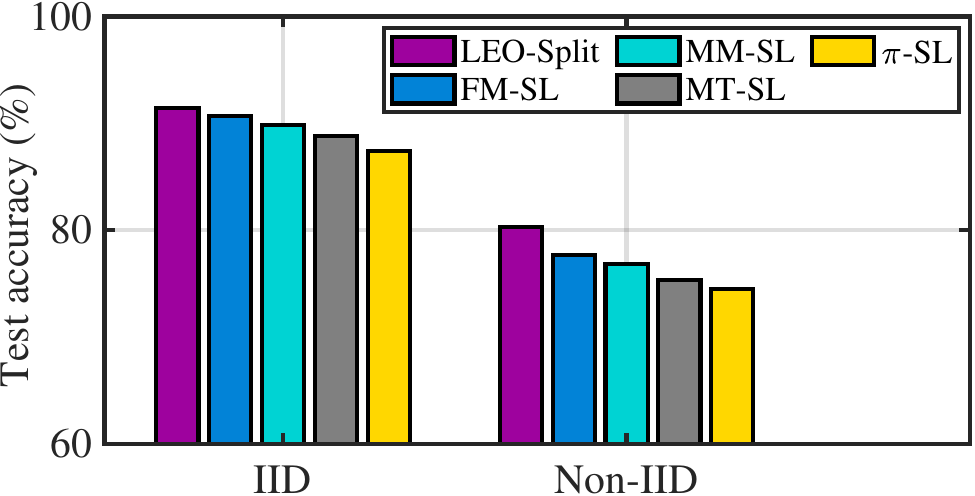}
  }
  \subfloat[Converged time (EuroSAT). \label{subfig:overall_bar_eus_conv}]
  {
    \includegraphics[width=0.476\linewidth]{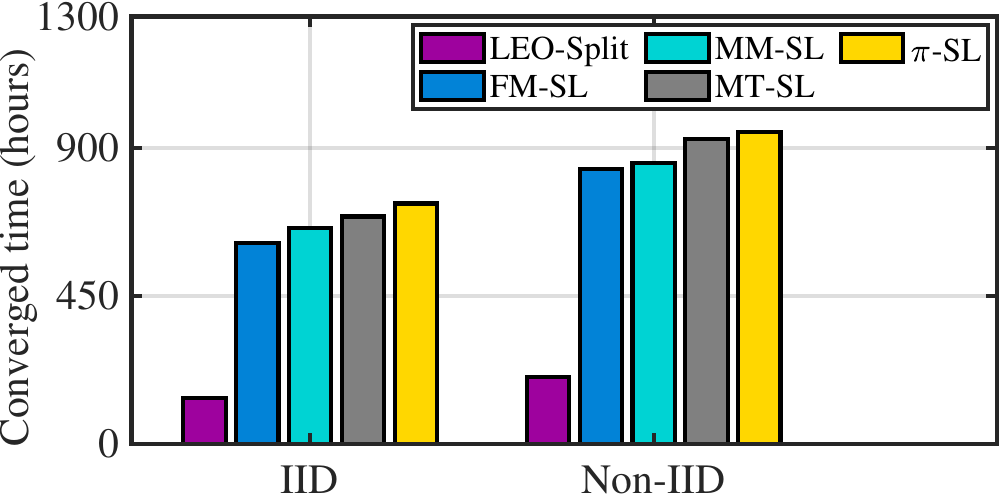}
  }
  \caption{Test accuracy and time on GBSense and EuroSAT under both IID and non-IID settings of VGG-16.}
  \label{fig:simulation_overall_bar}
  \vspace{-2ex}
\end{figure}

\subsection{Micro-benchmarking}
In this section, we investigate the impact of hyper-parameters on the model performance.

\subsubsection{The Impact of Labeling Rate}
Figure~\ref{fig:simulation_unlabel_impact} illustrates the impact of labeling rate on converged test accuracy on GBSense dataset under IID and non-IID settings. It is seen that \name and four benchmarks consistently exhibit worse performance with a lower labeling rate. \name experiences only a slight reduction in converged accuracy as the labeling rate drops from 10\% to 2\%. This is because \name leverages not only high-confidence pseudo-labeled data but also the rich information in low-confidence data that cannot generate pseudo-labels. Moreover, the design of PA in \name balances data class and quantity {across satellites} to enhance training performance, especially in the non-IID setting substantially. Conversely, the exclusive dependence of MT-SL and $\pi$-SL on feature-level consistency regularization significantly limits their ability to extract information from unlabeled data, leading to the severe accuracy deterioration, e.g., 5.1\% (10.4\%) and 5.5\% (11.7\%) accuracy loss {under IID (non-IID) setting}. For FM-SL and MM-SL, the lack of well-designed pseudo-label thresholds prevent them from adapting to varying learning status and difficulties across satellites, leading to relatively slight accuracy degradation compared to MT-SL and $\pi$-SL.

\begin{figure}[t!]
  \centering
  \subfloat[GBSense (IID).\label{subfig:simulation_unlabel_impact_iid}]{
    \includegraphics[width=0.472\linewidth]{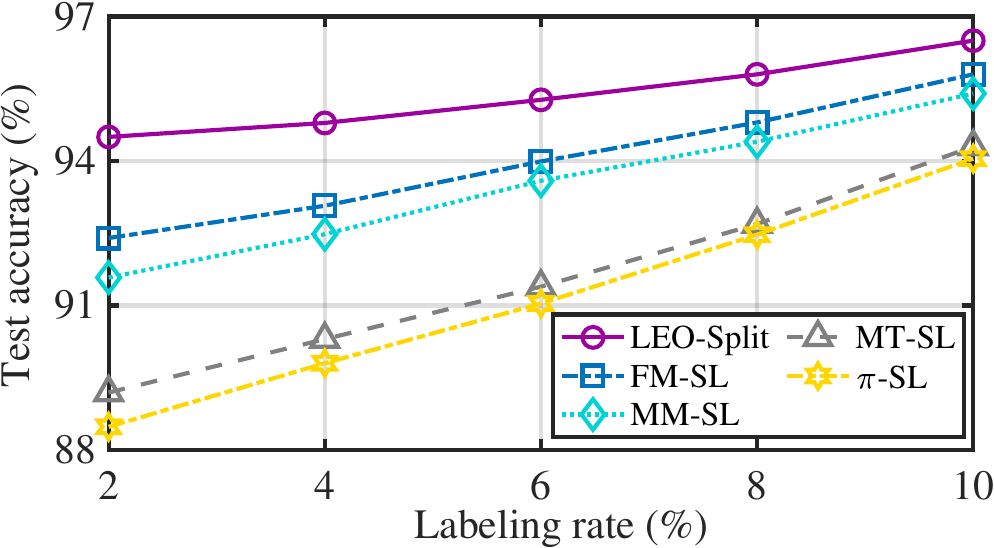}
  }
  \subfloat[GBSense (non-IID).\label{subfig:simulation_unlabel_impact_noniid}]
  {
    \includegraphics[width=0.472\linewidth]{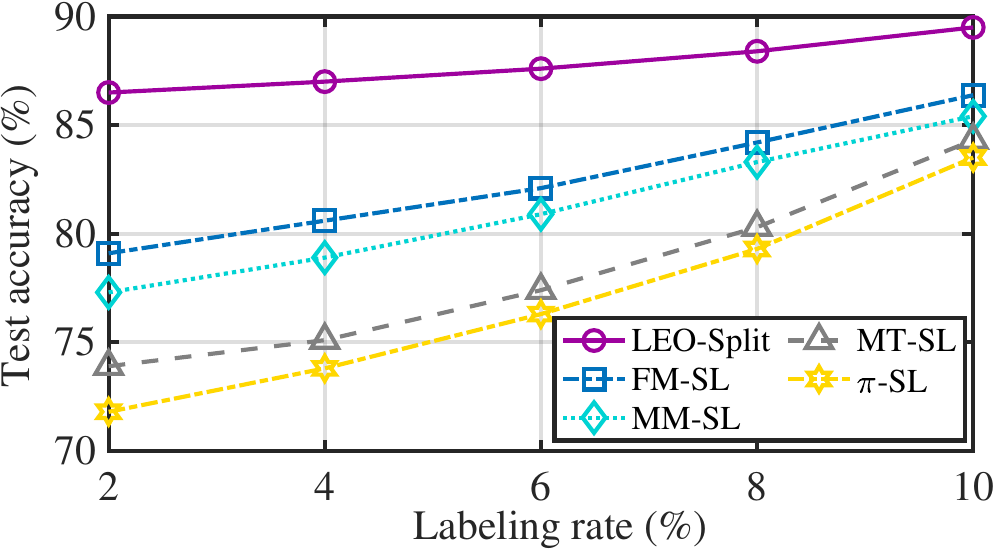}
  }
  \caption{The impact of labeling rate on test accuracy on GBSense dataset under IID and non-IID settings.}
  \label{fig:simulation_unlabel_impact}
\end{figure}

\begin{figure}[t]
  \centering
  \subfloat[{GBSense}.\label{subfig:simulation_heter_impact_gbsense}]{
    \includegraphics[width=0.472\linewidth]{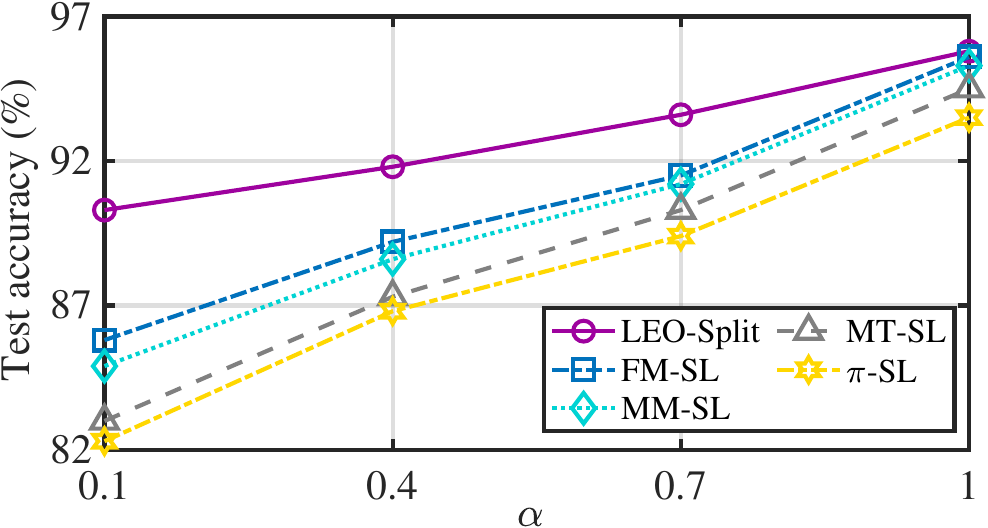}
  }
  \subfloat[EuroSATg.\label{subfig:simulation_heter_impact_eurosat}]
  {
    \includegraphics[width=0.472\linewidth]{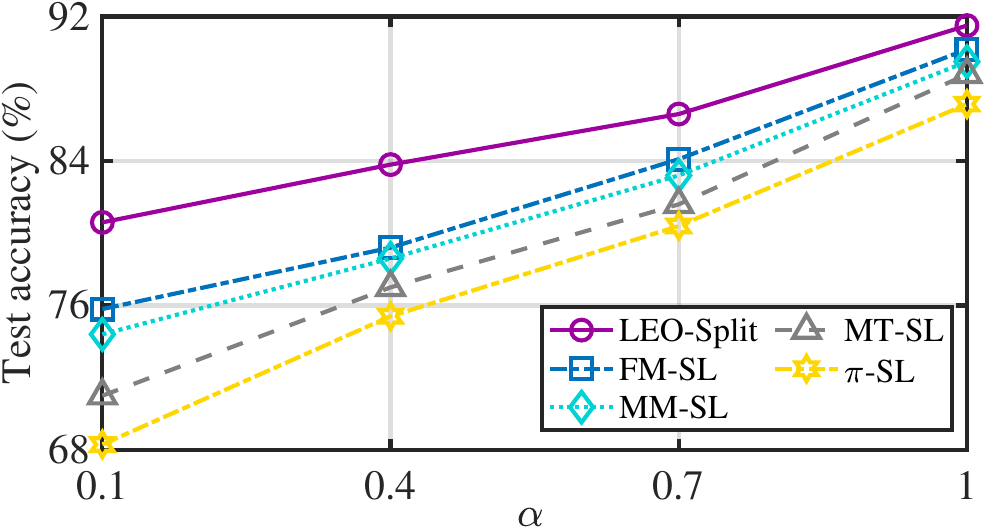}
  }
  \caption{The impact of data heterogeneity on converged test accuracy on GBSense and EuroSAT datasets.}
  \label{fig:simulation_heter_impact}
\end{figure}

\subsubsection{The Impact of Data Heterogeneity}
Figure~\ref{fig:simulation_heter_impact} illustrates the impact of data heterogeneity on the converged test accuracy on GBSense and EuroSAT datasets. As the level of data heterogeneity increases (indicated by a decrease in $\alpha$), the converged accuracy of \name and other benchmarks exhibits a significant decline. This drop in benchmarks is particularly pronounced, with an average accuracy loss of up to 11.2\% and 17.1\% on GBSense and EuroSAT datasets as $\alpha$ decreases from 1 to 0.1. The underlying reason for this deterioration is the lack of a balanced pseudo-labeling strategy for data class and quantity {across satellites}, which leads to significant local optimization biases in highly heterogeneous data and thus degrades model aggregation performance. In contrast, \name consistently outperforms other benchmarks, showcasing a minimal average accuracy loss of only 4.8\% and 9.9\% on GBSense and EuroSAT datasets. This is due to its threshold design in PA, where the class balance correction term enforces fewer pseudo-label generation for classes with more training data to balance discrepancies in {satellites} class distribution and the quantity balance correction term adaptively adjusts the relative ratio of thresholds across satellites to achieve {satellites} data quantity balance.

\begin{figure}[t!]
  \centering
  \subfloat[GBSense (IID).\label{subfig:simulation_contac_time}]{
    \includegraphics[width=0.472\linewidth]{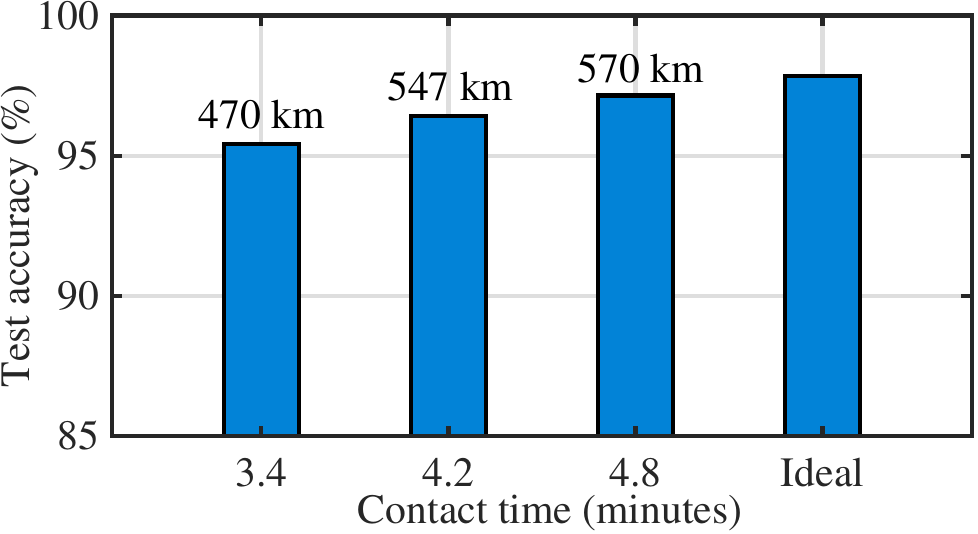}
  }
  \subfloat[GBSense (non-IID).\label{subfig:simulation_number_sate}]
  {
    \includegraphics[width=0.472\linewidth]{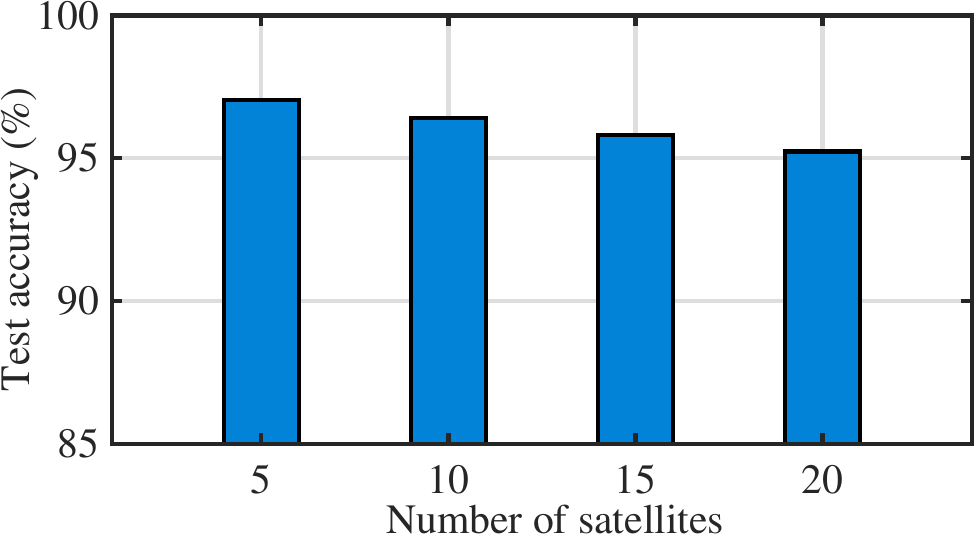}
  }
  \caption{The test accuracy versus contact time and the number of satellites on the GBSense dataset.}
  \label{fig:simulation_sat_para}
\end{figure}

\subsubsection{The Impact of Satellite Parameters}

%
We evaluate \name in the different contact time and number of satellites illustrated in Figures~\ref{subfig:simulation_contac_time} and~\ref{subfig:simulation_number_sate}, respectively. Figure~\ref{subfig:simulation_contac_time} shows that the performance of \name improves slightly with increasing contact time, but is close to the ideal case.  As suggested in Section~\ref{sssec:aux} and~\ref{ssec:server_side}, auxiliary model and adaptive activation interpolation jointly enable \name to largely maintain its performance under incomplete interactions.
Figure~\ref{subfig:simulation_number_sate}  shows that more participating satellites lead to a larger performance drop. The reason is that while the number of satellites increases, a satellite transmits fewer activations to a GS resulting in worse performance, given the limited downlink bandwidth.


\subsection{Ablation Study}
This section evaluates the impact of each module of \name on the model performance.

\subsubsection{Auxiliary Model (AM)}
Figure~\ref{subfig:simulation_ablation_localloss} shows the impact of AM on training performance. It is evident that PA+AAI converges approximately 4.5 times slower than \name, taking nearly 1652.3 hours to reach convergence. This marked slowdown occurs because excluding AM causes the loss of the satellite's ability to empower independent client-side sub-model update.
Consequently, limited contact time and transmission rate cause training failure during satellite-GS non-contact time, thus severely hindering the model convergence.
Moreover, \name exhibits comparable converged accuracy to PA+AAI, further showcasing the superior performance of AM.

\subsubsection{Pseudo-Labeling Algorithm (PA)}

Figure~\ref{subfig:simulation_ablation_class} and Figure~\ref{subfig:simulation_ablation_quantity}  illustrate the impact of class and quantity balance correction terms in PA on training performance. It is seen from Figure~\ref{subfig:simulation_ablation_class} that the removal of class balance correction term in the PA threshold (retaining the quantity balance correction term) results in a notable decline in training performance, with drops of 1.1\% at $\alpha=0.5$ and 1.8\% at $\alpha=0.2$. This is primarily attributed to catastrophic forgetting caused by discrepancies in data distribution among satellites and between satellites and GS. The class balance correction term enables classes with more training data to generate fewer pseudo-labels to balance the class distribution, thus achieving superior training performance. From Figure~\ref{subfig:simulation_ablation_quantity}, we see that the removal of quantity balance correction term results in accuracy loss of 1.4\% in mild (1:2:4) and 2.2\% in severe (1:5:10) quantity imbalance cases. The discrepancies in local dataset sizes among satellites tend to bias the updated model towards those satellites with more data samples. Conversely, the quantity balance correction term enables satellites with more training data to generate fewer pseudo-labels, compensating for the data quantity imbalance across satellites and thus improving converged accuracy.

\subsubsection{Adaptive Activation Interpolation (AAI)}

Figure~\ref{subfig:simulation_ablation_mixup} presents the impact of AAI on training performance. The removal of AAI results in an astounding cliff-like deterioration in training accuracy, plummeting from 89.5\% to 76.2\%, revealing the severe negative impact of server-side overfitting.
Additionally, the sparsity of activations inevitably causes data distribution inconsistencies between satellites and the GS, further degrading training performance. The design of AAI effectively enriches activations and balances their class distribution through linear interpolation, addressing the overfitting issue while guaranteeing data distribution between satellites and the GS.

\begin{figure}[t]
  \centering
  \subfloat[AM. \label{subfig:simulation_ablation_localloss}]
  {
    \includegraphics[width=0.470\linewidth]{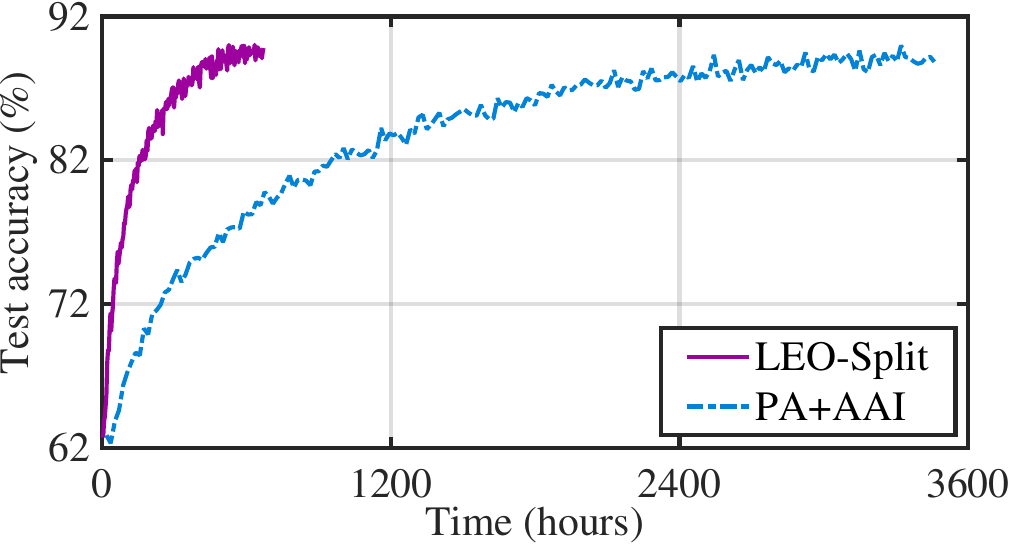}
  }
    \subfloat[PA (class). \label{subfig:simulation_ablation_class}]
  {
    \includegraphics[width=0.470\linewidth]{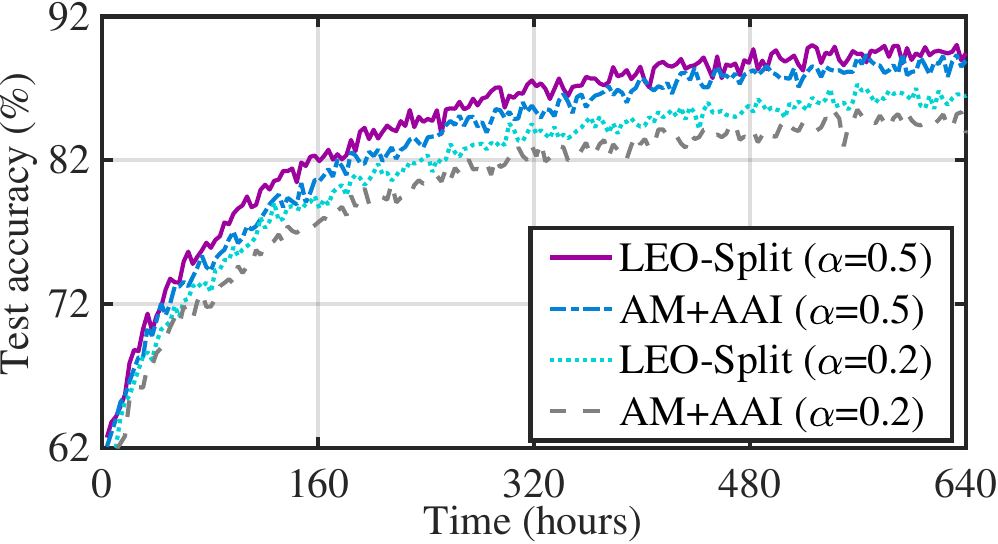}
  } \\
  \subfloat[PA (quantity).
  \label{subfig:simulation_ablation_quantity}]
  {
    \includegraphics[width=0.470\linewidth]{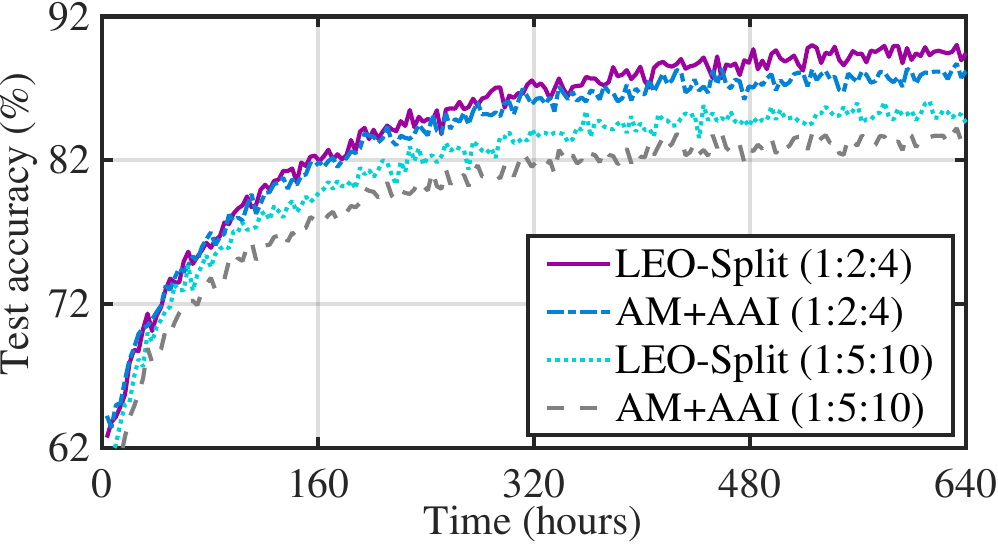}
  }
  \subfloat[AAI. \label{subfig:simulation_ablation_mixup}]
  {
    \includegraphics[width=0.470\linewidth]{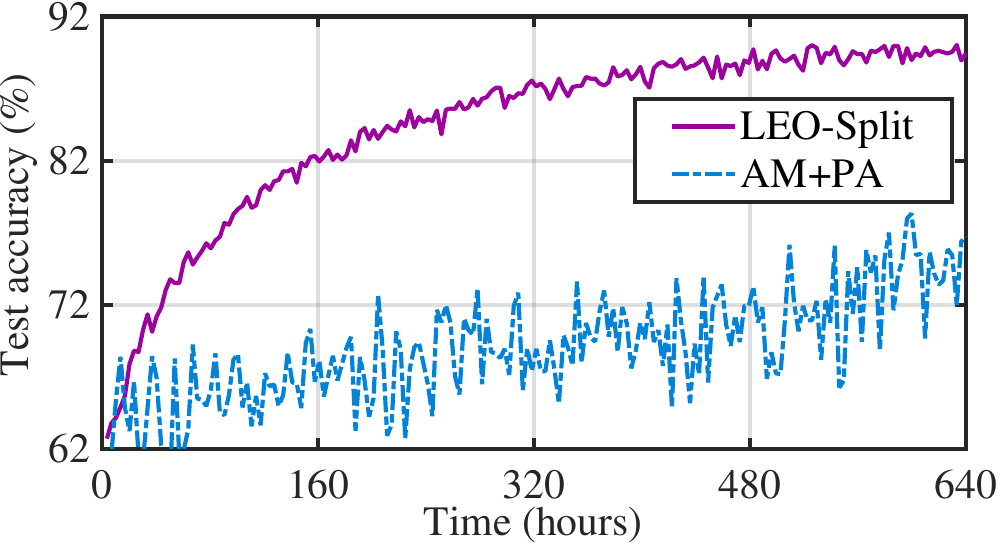}
  }
  \caption{Ablation experiments for AM (a), PA (class) (b), PA (quantity) (c), and AAI (d) on the GBsense dataset under non-IID settings.}
  \label{fig:simulation_ablation}
  \vspace{-2ex}
\end{figure}

\section{Related Work and Discussion}\label{sec:related_work}

\paragraph{Distributed Learning over Satellite Networks} Most research on deploying distributed learning over satellite networks exclusively focuses on FL~\cite{elmahallawy2024communication,elmahallawy2022fedhap,zhai2023fedleo,so2022fedspace} and is still nascent.
Elmahallawy~\textit{et al.}~\cite{elmahallawy2024communication,elmahallawy2022fedhap} propose novel FL frameworks tailored for LEO satellite constellations, utilizing high-altitude platforms as
distributed parameter servers to facilitate FL model training. So~\textit{et al.}~\cite{so2022fedspace} develop an innovative FL framework based on the trade-off between idle connectivity and local model staleness in satellite orbits. Zhai~\textit{et al.}~\cite{zhai2023fedleo} devise an offloading-assisted decentralized FL framework in LEO satellite networks,  which introduces additional iterative process in the offloading policy based on the training property of FL, thus mitigating the straggler effect in decentralized FL.
However, as ML models scale up, the FL paradigm faces significant challenges in deploying on resource-limited LEO satellites. Although SL can overcome the weakness of FL, research on SL in satellite networks is still void.



\paragraph{Semi-supervised Learning} Semi-supervised learning (SSL) is a paradigm for training models with limited labeled data and much more unlabeled data. There are two major research directions: One is consistency regularization~\cite{tarvainen2017mean,laine2016temporal,kuo2020featmatch}. {Tarvainen~\textit{et al.}~\cite{tarvainen2017mean} develop a student-teacher training framework, where the loss relies on output consistency between the teacher and student models, and the teacher model is an exponential moving average of the student model.}
Laine~\textit{et al.}~\cite{laine2016temporal} employ diverse data augmentation and dropout techniques to generate self-supervised signals, and utilize Euclidean distance between outputs from different network branches for guiding model training.
Kuo~\textit{et al.}~\cite{kuo2020featmatch} propose a novel feature-based refinement and augmentation method that generates diverse complex transformations for consistency-based regularization loss by utilizing prototypical representations extracted through clustering.
The other focus on pseudo-labeling~\cite{liu2022acpl,sohn2020fixmatch,berthelot2019mixmatch,berthelot2020remixmatch}. Sohn~\textit{et al.}~\cite{sohn2020fixmatch} employ the pre-defined fixed threshold to generate high-confidence pseudo-labels on weakly augmented unlabeled data for guiding model training.
{Liu~\textit{et al.}~\cite{liu2022acpl} design a new mechanism to select highly informative unlabelled samples for pseudo labeling
and an ensemble of classifiers to produce accurate pseudo-labels.}
{Berthelot~\textit{et al.}~\cite{berthelot2019mixmatch,berthelot2020remixmatch} enriches the unlabeled data by linear interpolation to generate high-confidence pseudo-labels, further improving training performance.}
While achieving significant progress, these methods solely rely on pre-defined fixed thresholds for pseudo-labeling, which may not well adapted to the class and quantity imbalances caused by the extreme heterogeneity of satellite orbital circumstances and hardware capabilities.

\paragraph{Discussion} There are several works~\cite{vasisht2021l2d2,10.1145/3570361.3592521} that study quickly downloading data from LEO satellites using distributed GSs built by low-cost commodity
hardware instead of highly specialized one~(i.e., multi-million US dollars). However, to maintain regulatory restrictions on different districts,
commodity GSs are not allowed to receive data that does not belong to them, thus preventing distributed GS systems from properly working. We believe SL is an alternative solution providing a new avenue for onboard data processing.

\section{Conclusion}\label{sec:conclusion}

Taking an important step towards satellite artificial intelligence, we have proposed a cutting-edge semi-supervised SL framework tailored for LEO satellite networks, named \name, aimed at enhancing model training effectiveness. \name consists of three primary components: {auxiliary model, pseudo-labeling algorithm, and adaptive activation interpolation}. First, the auxiliary model construction enables independent updating of client-side sub-models, resolving the training failure of the satellite-GS non-contact time. Second, the pseudo-labeling algorithm customizes adaptive pseudo-labeling threshold to mitigate catastrophic forgetting and model bias caused by data class and quantity imbalance. Lastly, the adaptive activation interpolation enriches server-side sub-model training data at GS by activation interpolation to prevent overfitting. Extensive experiments with real-world LEO satellite {traces} demonstrate that our \name framework achieves superior performance compared to state-of-the-art benchmarks. As a potential future direction, we are looking forward to extending our \name to improve the performance of various applications such as large language models~\cite{lin2024splitlora,hu2024agentscodriver,fang2024automated,lin2023pushing,hu2024agentscomerge}, multi-modal training~\cite{zheng2023autofed,hu2024t,fang2024ic3m}, etc in LEO satellite networks.

\ifCLASSOPTIONcaptionsoff
  \newpage
\fi

\bibliographystyle{IEEEtran}
\bibliography{reference}

\end{document}